\documentclass{article}

    \PassOptionsToPackage{numbers, compress}{natbib}

    \usepackage[final]{neurips_2022}

\usepackage[utf8]{inputenc} 
\usepackage[T1]{fontenc}    
\usepackage{hyperref}       
\usepackage{url}            
\usepackage{booktabs}       
\usepackage{amsfonts}       
\usepackage{nicefrac}       
\usepackage{microtype}      

\usepackage[dvipsnames]{xcolor}

\usepackage{graphicx}
\usepackage{subcaption}
\usepackage[most]{tcolorbox}
\usepackage{multirow}
\usepackage{todonotes}
\usepackage{float}
\usepackage{enumitem}
\usepackage{xhfill}
\usepackage{wrapfig}

\RequirePackage{fancyhdr}
\RequirePackage{color}
\RequirePackage{algorithm}
\RequirePackage{algorithmic}
\RequirePackage{natbib}
\RequirePackage{eso-pic} 
\RequirePackage{forloop}

\usepackage{amsmath,amsthm,amsfonts,bm,amssymb,mathrsfs}
\theoremstyle{plain}

\def\eqref#1{equation~\ref{#1}}

\def\1{\bm{1}}

\def\vg{{\bm{g}}}

\def\vx{{\bm{x}}}

\DeclareMathAlphabet{\mathsfit}{\encodingdefault}{\sfdefault}{m}{sl}
\SetMathAlphabet{\mathsfit}{bold}{\encodingdefault}{\sfdefault}{bx}{n}

\def\gB{{\mathcal{B}}}

\def\gD{{\mathcal{D}}}

\def\gG{{\mathcal{G}}}

\def\gL{{\mathcal{L}}}

\newcommand{\R}{\mathbb{R}}

\DeclareMathOperator*{\argmax}{arg\,max}

\renewcommand{\vec}[1]{\boldsymbol{#1}}

\newcommand{\e}{{\boldsymbol{\epsilon}}}

\newcommand\params{\boldsymbol{\theta}} 
\newcommand\basin{\boldsymbol{\Omega}} 
\newcommand\mhe{\boldsymbol \lambda_{\operatorname{max}}} 

\newcommand{\SWASAM}{\text{WASAM}} 
\newcommand{\SAM}{\text{SAM}} 
\newcommand{\SWA}{\text{SWA}} 
\newcommand{\NF}{\text{NF}} 

\newcommand{\ehat}{{\widehat{\vec{\epsilon}}}} 

\newcommand\loss{\gL}

\newcommand\nfsol{\params^{\text{NF}}}
\newcommand\samsol{\params^{\text{SAM}}}
\newcommand\swasol{\params^{\text{SWA}}}

\newcommand\wasamsol{\params^{\text{SWA+SAM}}}
\newcommand\cka{s_{\text{CKA}}}
\newcommand\cossim{s_{\text{cosine}}}

\newcommand\trainloss{\gL_{\text{train}}}

\newcommand\best[1]{\mathbf{#1}}
\newcommand\good[1]{\textcolor{ForestGreen}{#1}}
\newcommand\bad[1]{\textcolor{red}{#1}}

\definecolor{norange}{RGB}{249, 146, 0}
\definecolor{nred}{RGB}{234, 47, 32}
\definecolor{nblue}{RGB}{0, 100, 255}
\definecolor{npurple}{RGB}{146, 26, 192}
\usepackage{hyperref} 
\usepackage{cleveref}
\Crefname{section}{Sec.}{Secs.}
\Crefname{equation}{Eq.}{Eqs.}
\Crefname{figure}{Fig.}{Figs.}
\Crefname{tabular}{Tab.}{Tabs.}
\usepackage{soul}

\title{When Do Flat Minima Optimizers Work?}

\author{%
Jean Kaddour\thanks{Equal contribution, correspondence to \{jean.kaddour,linqing.liu\}.20@ucl.ac.uk}\\
Centre for Artificial Intelligence\\
University College London
\And
Linqing Liu$^*$\\
Centre for Artificial Intelligence\\
University College London
\And
Ricardo Silva\\
Department of Statistical Science\\
University College London
\And
Matt J. Kusner\\
Centre for Artificial Intelligence\\
University College London
}

\begin{document}

\maketitle

\begin{abstract}
Recently, \emph{flat-minima optimizers}, which seek to find parameters in low-loss neighborhoods, have been shown to improve a neural network's generalization performance over stochastic and adaptive gradient-based optimizers. Two methods have received significant attention due to their scalability: 1. Stochastic Weight Averaging (SWA), and 2. Sharpness-Aware Minimization (SAM). However, there has been limited investigation into their properties and no systematic benchmarking of them across different domains. 
We fill this gap here by comparing the loss surfaces of the models trained with each method and through broad benchmarking across computer vision, natural language processing, and graph representation learning tasks. We discover several surprising findings from these results, which we hope will help researchers further improve deep learning optimizers, and practitioners identify the right optimizer for their problem.
\end{abstract}

\section{Introduction} \label{introduction}
\begin{wrapfigure}{r}{0.35\textwidth}
    \centering
    \includegraphics[width=0.35\textwidth]{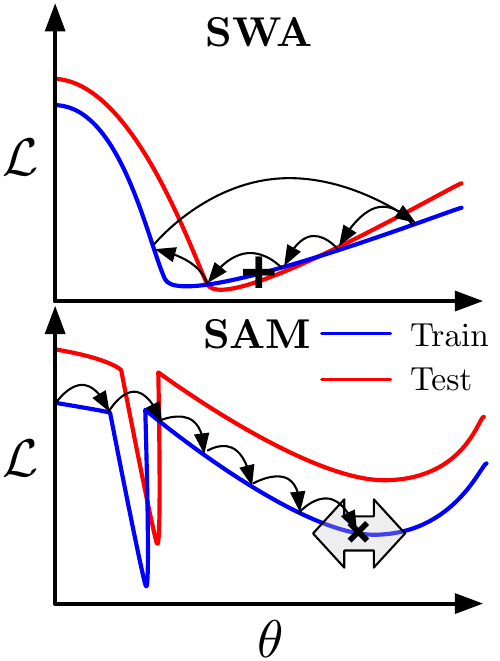}
    \caption{The mechanics behind SWA and SAM, whose solution is denoted by $\bm{+}$ and $\bm{\times}$, respectively. SWA produces a solution $\theta$ that is pulled towards flatter regions, while SAM approximates sharpness within the parameters' neighborhood (arrows). \vspace{-5ex}}
    \label{fig:teaser}
\end{wrapfigure}

Stochastic gradient descent (SGD) methods are central to neural network optimization \citep{bottou2018optimization}. 
Recently, one class of algorithms has focused on biasing SGD methods towards so-called `\emph{flat}' minima, which are located in large weight space regions with very similar low loss values \cite{hochreiter1997flat}. 
Theoretical and empirical studies \cite{PACBayes, relative_flatness, ESGD, BATCH, fantastic,low_pass_filter,SAM_VIT} postulate that such flatter regions generalize better than sharper minima, e.g., due to the flat minimizer's robustness against loss function shifts between train and test data, as illustrated in \Cref{fig:teaser}. Two popular flat-minima optimization approaches are: 1. Stochastic Weight Averaging (SWA) \cite{SWA}, and 2. Sharpness-Aware Minimization (SAM) \cite{SAM}.

While both strategies aim to find flatter minima, they operate much differently. 
On the one hand, SWA is based on the intuition that, near a flat minimum, gradients are smaller, leaving many iterates in that flat region. Therefore, averaging iterates will produce a solution that is pulled towards these flatter regions, see \Cref{fig:teaser}, top. 
On the other hand, SAM minimizes the maximum loss around a neighborhood of the current iterate. This way, a region around the iterate is designed to have uniformly low loss; see \Cref{fig:teaser}, bottom. Crucially, SAM requires an additional forward/backward pass for each parameter update, making it more expensive than SWA. 

Despite the successes \cite{SWA_RL, SWA_SSL, lawa, SAM_VIT, SAM_NLP} of SWA and SAM in some domains, we are unaware of a systematic comparison between them that would help practitioners to choose the right optimizer for their problem and researchers to develop better optimizers. The SWA \cite{SWA} paper was published in 2018, and the SAM \cite{SAM} paper in 2021; however, the SAM paper, and its most noticeable follow-ups \cite{ASAM, SAM_VIT, GSAM}, do not compare against SWA. Further, there is very limited overlap in terms of the model architecture and dataset used in the experiments among both papers, which are likely further confounded by other differences in the training procedures (e.g. data augmentations, hyper-parameters, etc.). 

\textbf{Contributions}
\begin{enumerate}[leftmargin=*]
    \item \textbf{In-depth comparison of minima found by SWA and SAM:} We visualize linear interpolations between different models and quantify the minimizers' flatnesses. This analysis yields 4 insights, e.g., despite SAM finding flatter solutions than SWA as quantified by Hessian eigenvalues, they can be close to sharp directions, a phenomenon that has been overlooked in the previous SAM literature. Averaging SAM iterates leads to the flattest among all minima.     
    \item \textbf{Rigorous comparison of SWA and SAM's performance over 42 tasks:} We empirically compare the optimizers with a rigorous model selection procedure on a broad range of tasks across different domains (CV, NLP, and GRL), model types (MLPs, CNNs, Transformers) and tasks (classification, self-supervised learning, open-domain question answering, natural language understanding, and node/graph/link property prediction). We discuss 9 findings, e.g., that both dataset and architecture impact their effectiveness, that for NLP tasks, SAM improves over SWA in most cases, and that the converse holds for GRL tasks. When flat-minima optimizers do not help, we notice clear discrepancies between the shapes of loss and accuracy curves. To assist future work, we open-source the code for all pipelines and hyper-parameters to reproduce the results. 
\end{enumerate}

\section{Background and Related Work} \label{sec:background}
\subsection{Stochastic Gradient Descent (SGD)}
The classic optimization framework of machine learning is empirical risk minimization 
\begin{align}
\gL\left(\params\right)=\frac{1}{N} \sum_{i=1}^{N} \ell\left(\vx_{i}; \params\right)
\end{align} where $\params \in \mathbb{R}^{d}$ is a vector of parameters, $\left\{\vx_{1}, \ldots, \vx_{N}\right\}$ is a training set of inputs $\vx_n \in \R^D$, and $\ell(\vx; \params)$ is a loss function quantifying the performance of parameters $\params$ on  $\vx$. SGD samples a minibatch $\mathcal{S} \subset\{1, \ldots, N\}$ of size $|\mathcal{S}| \ll N$ from the training set and updates the parameters through 
\begin{align}
    \params^{\text{SGD}}_{t+1} = \params_t - \eta \vg \left(\params_t\right), \; \text{where} \; \vg \left(\params\right)=\frac{1}{|\mathcal{B}|} \sum_{i \in \mathcal{B}} \nabla \ell\left(\params; \vx_{i}\right), \label{eq:sgd_update}
\end{align} 
for a length specified by $\eta$, the learning rate. 

\begin{figure} \centering
\begin{minipage}{0.49\textwidth}
\begin{algorithm}[H]
   \caption{Stochastic Weight Averaging \cite{SWA}} 
   \label{alg:swa}
\begin{algorithmic}
   \STATE {\bfseries Input:} Loss function $\loss$, training budget in number of iterations $b$, training dataset $\gD:=\cup^n_{i=1}\{\vx_i\}$, mini-batch size $|\gB|$, averaging start epoch $E$, averaging frequency $\nu$, (scheduled) learning rate $\eta$, initial weights $\params_0$.
   \FOR{ $k \leftarrow 1, \dots, b$} 
   \STATE \hspace{-0.2cm} Sample a mini-batch $\gB$ from $\gD$
    \STATE \hspace{-0.2cm} Compute gradient $\bm g \leftarrow \nabla \loss\left(\params_t\right)$
   \STATE \hspace{-0.2cm} Update parameters $\params_{t+1} \leftarrow \params_{t} - \eta \bm g$
   \IF {$k \geq E$ and ${\bmod(k, \nu) = 0}$} 
    \STATE ${\swasol_{t+1} = \left(\swasol_t \cdot l + \swasol_{t+1}\right) / \left(l + 1\right)}$
   \ENDIF
   \ENDFOR
   \STATE {\bfseries return} $\swasol$
\end{algorithmic}
\end{algorithm}
\end{minipage}
\begin{minipage}{0.49\textwidth}
\begin{algorithm}[H]
   \caption{Sharpness-Aware Minimization \cite{SAM}} 
   \label{alg:sam}
\begin{algorithmic}
   \STATE {\bfseries Input:} Loss function $\loss$, training budget in number of iterations $b$, training dataset $\gD:=\cup^n_{i=1}\{\vx_i\}$, mini-batch size $|\gB|$, neighborhood radius $\rho$, (scheduled) learning rate $\eta$, initial weights $\params_0$.
   \FOR{ $k \leftarrow 1, \dots, b$} 
   \STATE \hspace{-0.2cm} Sample a mini-batch $\gB$ from $\gD$
   \STATE \hspace{-0.2cm} Compute worst-case perturbation ${\ehat \leftarrow \rho \frac {\displaystyle \nabla \loss(\params)} {\displaystyle \Vert \nabla \loss(\params) \Vert_2}}$ 
    \STATE \hspace{-0.2cm} Compute gradient $\bm g \leftarrow \nabla \loss\left(\samsol_t {+ \ehat}\right)$
   \STATE \hspace{-0.2cm} Update parameters $\samsol_{t+1} \leftarrow \samsol_{t} - \eta \bm g$
   \ENDFOR
   \STATE {\bfseries return} $\samsol$
\end{algorithmic}
\end{algorithm}
\end{minipage}
\end{figure}

\subsection{Stochastic Weight Averaging (SWA)} \label{sec:swa} The idea of averaging weights dates back to accelerating the convergence speed of SGD \cite{polyak_avg,lawa}. 
SWA's motivation is based on the following observation about SGD's behavior when training neural networks: it often traverses regions of the weight space that correspond to high-performing models but rarely reaches the central points of this optimal set. Averaging the parameter values over iterations moves the solution closer to the centroid of this space of points. 

The SWA update rule is the cumulative moving average \begin{align}
\params^{\text {SWA }}_{t+1} \leftarrow \frac{\params_{t}^{\text {SWA }} \cdot l+\params_t^{\text{SGD}}}{l+1},
\end{align}where $l$ is the number of distinct parameters averaged so far and $t$ is the SGD iteration number.\footnote{SWA parameters are constant between averaging steps.} 

SWA has two hyper-parameters: the update frequency $\nu$ and starting epoch $E$. When using a constant learning rate, \citet{SWA} suggests updating the parameters once after each epoch, i.e., $\nu \approx \frac{N}{|\mathcal{B}|}$, and starting at $E \approx 0.75 T$, where $T$ is the training budget required to train the model until convergence with conventional SGD training. 

\citet{AV} argue that SWA may always improve generalization, regardless of the loss function's geometry. \citet{lawa} show that averaging a specific range of weights can speed up training convergence. \citet{SWAD} argue that tuning $\nu$ and $E$ carefully is necessary to make it work effectively in domain generalization (DG) tasks. Besides DG tasks, a list of tuned hyper-parameters based on a fair model selection procedure across different architectures and tasks has been missing in the literature. To the best of our knowledge, \citet{SWAD} is the only study that compares SWA and SAM over the same experiments, but it focuses on domain generalization tasks which we, therefore, leave out in this work. 

\subsection{Sharpness-Aware Minimization (SAM)} \label{sec:sam}
While SWA is implicitly biased towards flat minima, SAM \emph{explicitly} approximates the flatness around parameters $\params$ to guide the parameter update. It first computes the worst-case perturbation $\e$ that maximizes the loss within a given neighborhood $\rho$, then minimizes the loss w.r.t. the perturbed weights $\params + \e$. Formally, SAM finds $\params$ by solving the minimax problem:
\begin{align}
\label{eq:sam-loss}
\min_{\params} \max_{||\e||_2\leq \rho} \loss(\params+\e),
\end{align}
where $\rho \geq 0$ is a hyperparameter.

To find the worst-case perturbation $\e^*$ efficiently in practice, \citet{SAM} approximates \Cref{eq:sam-loss} via a first-order Taylor expansion of $\loss(\params+\boldsymbol{\epsilon})$ w.r.t. $\boldsymbol{\epsilon}$ around $\boldsymbol{0}$, obtaining
\begin{align}
\label{eq:fo_approximation}
    {\boldsymbol{\epsilon}^*} & \approx  \argmax_{\|\boldsymbol{\epsilon}\|_2 \leq \rho} \boldsymbol{\epsilon}^{\top} \nabla_{\params} \loss(\params) \approx \underbrace{\rho \cdot \frac{\nabla_{\params} \loss (\params)}{\left \| \nabla_{\params} \loss(\params) \right \|}}_{=: \ehat}.
\end{align} In words, $\ehat$ is simply the scaled gradient of the loss function w.r.t to the current parameters $\params$. Given $\ehat$, the altered gradient used to update the current $\params_t$ (in place of $\vg(\params_t)$) is
\begin{equation*}
\label{eq:sam-gradient}
\nabla_{\params}  \max_{||\boldsymbol{\epsilon}||_2\leq \rho} \loss(\params+\boldsymbol{\epsilon}) \approx {\nabla_{\params} \loss(\params)|_{\params+\hat{\boldsymbol{\epsilon}}}}.
\end{equation*}

Due to \Cref{eq:fo_approximation}, SAM's computational overhead consists of an additional forward and backward pass per parameter update step compared to SWA and non-flat optimizers.

SAM's performance strongly depends on the neighborhood radius $\rho$. For example, \citet{SAM_VIT, wu2022adversarial} show that $\rho$ should be set to values outside the originally considered ranges by \citet{SAM}. Analogously to \Cref{sec:swa}, this lack of coherence among hyper-parameter tuning protocols in the SAM literature makes it tricky to determine SAM's comparative effectiveness.

\subsection{Other Flat-Minima Optimizers}
There are several extensions of SWA \cite{SWA_REVISITED, SWAD} and SAM \cite{ASAM, GSAM, zhao2022ss}. For simplicity, we do not consider them in this work. Besides SWA and SAM, other flat-minima optimizers include e.g., \cite{ESGD, sankar2020deeper}. However, due to their computational cost and/or lack of performance gains, we do not include them in this work. \citet{ESGD} requires $[5, 20]$ forward and backward passes per parameter update. \citet{sankar2020deeper} similarly requires $[5, 10]$ forward and backward passes to estimate the Hessian trace and $6$ of $7$ experiments yield minimal improvement of $\leq 0.27\%$, see Table 1 in \citet{sankar2020deeper}. In contrast, SWA and SAM have been shown to increase performance by multiple percentage points in some cases \cite{SWAD, SAM_VIT} while requiring fewer computational resources. 
\section{How do minima found by SWA and SAM differ?} \label{sec:main}
In this section, we investigate SWA and SAM solutions in two prototypical deep learning tasks, where these optimizers improve over the baseline. Our goal is to understand better their geometric properties (instead of their generalization performance, which is the focus of \Cref{sec:benchmark}). 

First, we investigate the behavior of the loss landscape along the line between non-flat and flat solutions (\Cref{subsec:interpolations}). Previous studies successfully used such linear interpolations to gain novel insights, e.g., for training dynamics \cite{linear_path_gf, revisiting_linear_path}, regularization \cite{visualizing_loss_surfaces, stochasticity}, and network pruning \cite{LMC_LTH}. Second, motivated by findings in \Cref{subsec:interpolations}, we average SAM iterates and visualize interpolations between averaged and non-averaged solutions (\Cref{subsec:wasam}). Interestingly, the averaged SAM solution is less susceptible to asymmetric directions. Third, we compare quantitative measurements of all solutions' flatnesses (\Cref{subsec:quantitative}). Here, we compute dominant Hessian eigenvalues, as commonly used in the flat minima literature \cite{ESGD, pyhessian, SAM_VIT, SAM}. Lastly, in \Cref{app:cka}, we further compute CKA \cite{cka_similarity} and cosine similarities between SWA/SAM's network output logits. 

We choose the following two disparate learning settings: (i) a well-known image classification task, widely used for evaluation in flat-minima optimizer papers, and (ii) a novel, challenging Python code summarization task, on which state-of-the-art models achieve only around $16\%$ F1 score on the test set (which is \textbf{higher} than its commonly achieved accuracy on the more challenging training set), and that has not been explored yet in the flat-minima literature. Specifically, for (i), we investigate the loss/accuracy surfaces of a WideResNet28-10 \cite{wrn} model on CIFAR-100 \cite{cifar} (baseline non-flat optimizer: SGD with momentum (SGD-M)) \cite{momentum}. For (ii), we use the theoretically-grounded Graph Isomorphism Network \cite{GIN} model on OGB-Code2 \cite{ogb} (baseline optimizer: Adam \cite{adam}).

All optimizers start from the same initialization. We denote the minimizer produced by the non-flat methods (SGD-M and Adam) by $\nfsol$ and the flat ones by $\swasol$ and $\samsol$. 

\subsection{What is between non-flat and flat solutions?}  \label{subsec:interpolations} We start by comparing the similarity of flat and non-flat minimizers through linear interpolations. This analysis allows us to understand if they are in the same basin and how close they are to a region of sharply-increasing loss, where we expect loss/accuracy to differ widely between train and test. Further, for each of our four observations, we recommend a future work direction.

To linearly interpolate between two sets of parameters $\params$ and $\params^{\prime}$, we parameterize the line connecting these two by choosing a scalar parameter $\alpha$ and defining the weighted average $\params(\alpha) = (1-\alpha)\params + \alpha \params^{\prime}$. If there exists no high-loss barrier between two networks $\params, \params^{\prime}$ along the linear interpolation, we say that they are located in the same \emph{basin}, i.e., $\{\params, \params^{\prime} \} \in \basin$. \cite{transfer_learning,zhou2020towards}. A basin is an area in the parameter space where the loss function has relatively low values. Due to NN non-linearities, the linear combination of the weights of two accurate models does not necessarily define an accurate model. Hence, we generally expect high-loss barriers along the linear interpolation path.

While there are alternative distance measures that could be used to compare two networks, they typically either (a) do not offer clear interpretations, as pointed out by \citet{LMC_LTH}, or (b) yield trivial network connectivity results, such as \emph{non-linear} low-loss paths, which can be found for any two network minimizers \cite{draxler, garipov2018loss, a_closer_look_at_dl_heuristics, neural_network_landscapes}.

\begin{figure*}[h!]
		\centering
		\includegraphics[width=\textwidth]{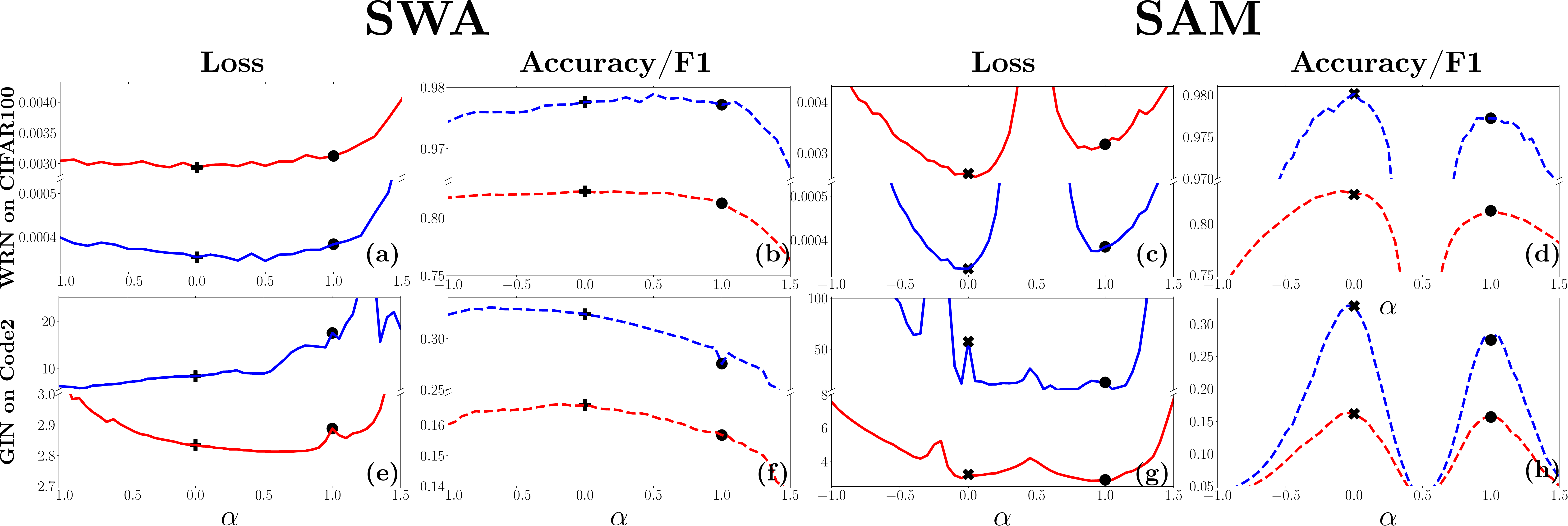}
		\caption{
		Training (\textcolor{blue}{blue}) and test (\textcolor{red}{red}) losses (---) and accuracies (\protect\makebox[1.3em]{\protect\xdotfill{.4pt}}) of linear interpolations $\params(\alpha)=(1-\alpha) \params+\alpha \params^{\prime}$ (for $\alpha \in [-1,1.5]$) between SWA ($\bm{+}$) and SAM ($\bm{\times}$) solutions ($\alpha=0.0$) and non-flat baseline solutions ($\protect \bullet, \alpha=1.0$). }
		\label{fig:flat_vs_non_flat}
	\end{figure*}

\textbf{Obs. 1: $\{\swasol, \nfsol\} \in \basin^{\text{NF}}$.}
$\swasol$ and $\nfsol$ are in the same basin, as can be seen in Figures~\ref{fig:flat_vs_non_flat}a and \ref{fig:flat_vs_non_flat}e. Additionally,  $\nfsol$ is near the periphery of a sharp increase in loss, as can be seen when moving in the direction from $\swasol$ to $\nfsol$ (i.e., $\alpha > 1$). Conversely, $\swasol$ finds flat regions that change slowly in the loss. This bias of SWA to flatter loss beneficially transfers to the accuracy landscape too: Figures~\ref{fig:flat_vs_non_flat}b and \ref{fig:flat_vs_non_flat}f show the accuracy/F1 score rapidly dropping off approaching and beyond $\nfsol$. Interestingly, in Figures~\ref{fig:flat_vs_non_flat}e and \ref{fig:flat_vs_non_flat}f, we see that for Code2, for $\alpha<0$, there exist solutions with even better training loss/accuracy but worse test loss/accuracy. However, $\swasol_{\text{GIN}}$ is close to the test accuracy maximizer along this interpolation. Future work may inspect why the cross entropy loss function used for GIN/Code2 seems less well correlated with its accuracy compared to WRN/CIFAR100.

\textbf{Obs. 2: $\samsol \in \basin^{\text{SAM}} \ne \basin^{\text{NF}}$.}
$\samsol$ and $\nfsol$ are not in the same basin: Figures~\ref{fig:flat_vs_non_flat}c and \ref{fig:flat_vs_non_flat}g show that there is a high loss barrier between them, respectively. Figures~\ref{fig:flat_vs_non_flat}d and \ref{fig:flat_vs_non_flat}h show that $\samsol$ and even nearby points in parameter space achieve higher accuracies/F1 scores (i.e., generalize better) than $\nfsol$ and points around it. This is an interesting result because we expect different basins to produce qualitatively different predictions, one of the motivations behind combining models, even if they exhibit different performances \cite{snapshot_ensembling, deep_ensembles}. \citet{grewaldiversity} successfully combine models yielded by different optimizers, and we think future work should study ensembling SAM and non-SAM solutions.

\textbf{Obs. 3: SAM finds a saddle point.}
Figure~\ref{fig:flat_vs_non_flat}g shows $\samsol_{\text{GIN}}$ being located in a sharp training loss minimum whose loss is much higher than $\nfsol$. Yet, its test loss is slightly higher, and its F1 score is better. We visualize 2D plots moving along random directions (not shown here due to space) to confirm that $\samsol_{\text{GIN}}$ is a saddle point (\Cref{app:sam_saddle}). A common pathology among curvature-based methods is that they attract saddle points \cite{saddle_points}. Since SAM takes some form of curvature into account, too, we believe that future work should investigate SAM's propensity to find saddle points and potential remedies.

\textbf{Obs. 4: $\samsol$ is closer to sharper directions than $\swasol$,}
as can be seen by $\gL_{\text{tr/te}}(\samsol(0.1)) \approx 2 \cdot \gL_{\text{tr/te}}(\samsol(-0.1))$, while $\gL_{\text{tr/te}}(\swasol(0.1)) \approx \gL_{\text{tr/te}}(\swasol(-0.1))$, where $\gL(\cdot)_{\text{tr/te}}$ refers to both training and test loss functions. A possible explanation for SAM being closer to sharp sides is that while it finds different basins than SGD/SWA by smoothing the loss surface (as illustrated in \Cref{fig:teaser}), \emph{within} a local basin, it may oscillate around the minimizer similarly as SGD. One cause for this can be that $\basin^{\text{SAM}}$'s hypersphere is larger than SAM's radius $\rho$. If that holds, then given a small enough learning rate, we expect it to oscillate around $\params^* \in \basin^{\text{SAM}}$ (the smaller the learning rate, the less likely it escapes the basin due to that stochasticity). Two possible remedies are: (1) adapt/schedule $\rho$, or (2) average SAM iterates to bias its solution towards the flatter side. (1) has been explored by \cite{GSAM,zhao2022ss}. We try (2) in the next subsection. Future work may study SAM's basin escape time, e.g., using convolutions \cite{kleinberg2018alternative} or stochastic differential equations \cite{zhou2020towards}.

\subsection{What happens if we average SAM iterates?} \label{subsec:wasam}
Based on observation 4: ``\textbf{$\samsol$ is closer to sharper directions than $\swasol$}'', averaging SAM iterates may further improve generalization, referred to as \emph{Weight-Averaged Sharpness-Aware Minimization} (WASAM) \footnote{A code implementation can be found here: \url{https://github.com/jeankaddour/WASAM}}. The reason is that while SAM finds better-performing basins, \emph{within} the basin, its final iterate may still be near a side that increases sharply in the loss.

\begin{figure*}[h!]
    \centering
    \includegraphics[width=\textwidth]{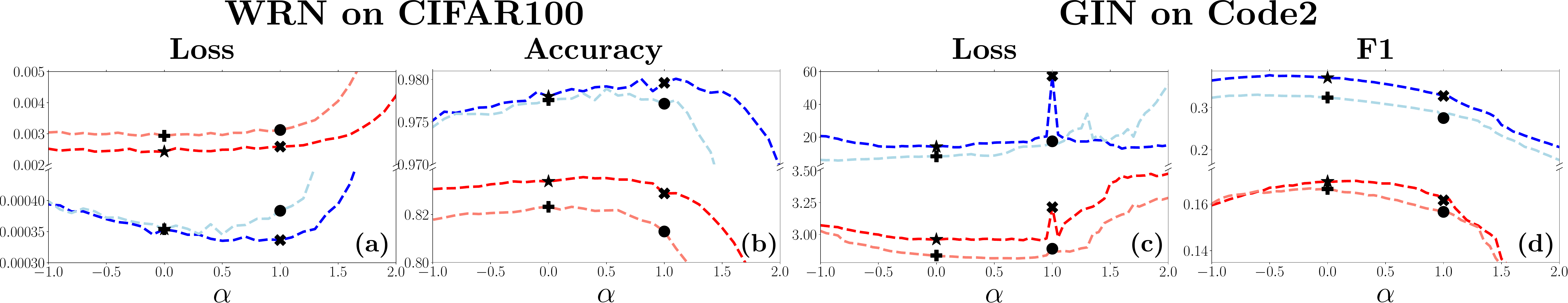}
    \caption{Training (\textcolor{blue}{blue}) / test (\textcolor{red}{red}) losses (---) / accuracies (\protect\makebox[1.3em]{\protect\xdotfill{.4pt}}) between \textcolor{gray}{non-flat baseline ($\bullet$) $ \leftrightarrow$ SWA ($\bm{+}$)}, SAM ($\bm{\times}$) $\leftrightarrow$ WASAM ($\star$).}
    \label{fig:wasam_works}
\end{figure*}

Starting with the first of the two previously analyzed settings (WRN/CIFAR100), Figures~\ref{fig:wasam_works}a, and \ref{fig:wasam_works}b show that $\wasamsol_{\text{WRN}}$ (marker: $\bigstar$) achieves the lowest test loss and highest test accuracy, respectively. What stands out in comparison to the previous plots is $\samsol_{\text{WRN}}$'s ($\bm{\times}$) proximity to sharp sides, surprisingly similar to $\nfsol_{\text{WRN}}$ ($\bullet$) here and in Figures~\ref{fig:flat_vs_non_flat}c and \ref{fig:flat_vs_non_flat}e. As we hoped, $\wasamsol_{\text{WRN}}$ is indeed closer to a flatter region, as can be seen by $\gL_{\text{tr/te}}(\wasamsol_{\text{WRN}}(-0.2)) \approx \gL_{\text{tr/te}}(\wasamsol_{\text{WRN}}(0.2))$.

In GIN/OGB-Code2, one unanticipated finding is that $\wasamsol_{\text{GIN}}$ escapes the (previously discussed) saddle point of $\samsol_{\text{GIN}}$, appearing here as a maximum in Figure~\ref{fig:wasam_works}c. A likely reason is that SAM traversed nearby flatter regions before arriving at the saddle point, especially if it is a non-strict saddle. In terms of F1 score, Figure~\ref{fig:wasam_works}d shows that while $\swasol_{\text{GIN}}$ ($\bm{+}$) and $\samsol_{\text{GIN}}$ perform about equally well, the flatter region found by $\wasamsol_{\text{GIN}}$ improves over both. 

\subsection{How ``flat'' are the found minima?} \label{subsec:quantitative}
We now quantify the flatnesses of all four optimizers over both tasks by computing the median of the dominant Hessian eigenvalue across all training set batches using the Power Iteration algorithm \cite{power_iteration, pyhessian}. This metric measures the worst-case loss landscape curvature. We choose this metric as it is very commonly used in the minima flatness literature, e.g., \cite{ESGD, SAM_VIT, SAM, yao2019large, hawq, physics_informed, at_flat_minima}.

\Cref{tab:hessian} shows that SAM leads to flatter minima than SWA in both cases. Interestingly $\mhe(\nfsol_{\text{WRN}}) \approx 2.5 \cdot \mhe(\{\swasol, \samsol \})$, while $\mhe(\nfsol_{\text{WRN}}) \approx 5.75 \mhe(\wasamsol_{\text{WRN}})$, indicating room for improvement in terms of flatness for both SWA and SAM. The relative differences are less dramatic for GIN/Code2, although surprisingly $\mhe(\nfsol_{\text{GIN}}) \approx  \mhe(\swasol_{\text{GIN}})$.
In sum, averaging SAM iterates leads to the flattest minima \textbf{and} best-performing minima in both cases (see \Cref{sec:benchmark}).
\begin{table}[h!]
    \centering
    \caption{Median $\boldsymbol \lambda_{\operatorname{max}}$ of Hessian over all training set batches.}
    \scalebox{1.}{
\begin{tabular}{c|c|c|c|c}
    \textbf{Task} & \textbf{Baseline} & \textbf{SWA} & \textbf{SAM} & \textbf{WASAM} \\  \hline 
     WRN on CIFAR100 & $673$ & $265 $ & $237$ & $ \bf 117$\\
     GIN on Code2 & $16.65$& $16.79$& $11.31$& $\bf9.96$ \\
    \end{tabular}}
    \label{tab:hessian}
    \vspace{-3ex}
\end{table}

\section{How do SWA and SAM perform on a broad set of experiments?}
\label{sec:benchmark}
As we point out in the introduction, there is almost no overlap and consistency regarding reported SWA and SAM results in the literature. This section addresses this gap. For example, \citet{SAM_NLP,SAM_VIT} illustrate that the flat minima found by SAM improve generalization on Transformer \cite{transformer} architectures compared to non-flat optimizers, but they do not compare against SWA. Hence, it is unclear if the computationally cheaper SWA may provide better or similar performance. 

We compare flat minimizers SWA, SAM, and averaged SWA iterates (WASAM) over the non-flat minimizers across a range of different tasks in the domains of computer vision, natural language processing, and graph representation learning. We average all runs at least three times across random seeds (more often for experiments with higher variability, see details in \Cref{app:experimental_details}), and we report the corresponding standard error.
We bold the best-performing approach and any approach whose average performance plus standard error overlaps it. 

\paragraph{Hyper-parameters.} \label{sec:hparams}
For all architectures and datasets, we set hyperparameters 
shared by all methods (e.g., learning rate) mostly to values cited in prior work \footnote{Sometimes with minor modifications, e.g., adjusting per-device batch sizes to be compatible with our GPU infrastructure.} As explained in \Cref{sec:swa,sec:sam}, the effectiveness of flat-minima optimizers is highly sensitive to their additional hyper-parameters. We select hyper-parameters using a grid search over a held-out validation set. Specifically, for SWA we follow \citet{SWA} and hold the update frequency $\nu$ constant to once per epoch and tune the start time $E \in \{0.5T, 0.6T, 0.75T, 0.9T \}$ ($T$ is the number of baseline training epochs). \citet{SWA} argue that a cyclical learning rate starting from $E$ helps to encourage exploration of the basin. For the sake of simplicity, we average the iterates of the baseline directly but include even earlier starting times (i.e., $0.5T, 0.6T$). In \Cref{app:constant_lr}, we compare against using a constant learning rate at the end of training. For SAM, we tune its neighborhood size $\rho \in \{0.01, 0.02, 0.05, 0.1, 0.2 \}$, as in previous work \cite{SAM, SAM_NLP}. 

\Cref{app:experimental_details} contains the values of all hyper-parameters and additional training details (including public model checkpoints, hardware infrastructure, software libraries, etc.) to ensure full reproducibility alongside open-sourcing our code. 

\begin{table}
\centering
\caption{CV test results: Supervised Classification (SC), and Self-Supervised Learning (SSL) tasks.}
\scalebox{0.7}{
\begin{tabular}{llcccccccc} \toprule
Task & Model & \multicolumn{1}{c}{ Baseline } & SWA & SAM &  WASAM\\ \midrule
\multirow{4}{*}{\shortstack[l]{SC:\\ CIFAR10}} & WRN-28-10 & $96.78_{\pm 0.03}$ &  \bad{$-0.05_{\pm 0.04}$} &  \good{ + $\best{0.34}_{\pm 0.09}$} &  \good{ + $0.25_{\pm 0.05}$} \\
& PN-272 & $96.73_{\pm 0.14}$ &  \good{ + $0.22_{\pm 0.14}$} &  \good{ + $\best{0.42}_{\pm 0.06}$} &  \good{ + $\best{0.41}_{\pm 0.02}$} \\
& ViT-B-16 &$98.95_{\pm 0.02}$ &  \bad{$-0.04_{\pm 0.04}$} &  \good{ + $0.07_{\pm 0.01}$} &  \good{ + $\best{0.10}_{\pm 0.01}$} \\
& Mixer-B-16 & $96.65_{\pm 0.03}$ &  \good{ + $0.02_{\pm 0.03}$} &  \good{ + $\best{0.19}_{\pm 0.05}$} &  \good{ + $\best{0.22}_{\pm 0.06}$}\\
\midrule
\multirow{4}{*}{\shortstack[l]{SC:\\ CIFAR100}} & WRN-28-10 & $80.93_{\pm 0.19}$ &  \good{ + $1.62_{\pm 0.06}$} &  \good{ + $1.82_{\pm 0.14}$} &  \good{ + $\best{2.24}_{\pm 0.14}$} \\
& PN-272 & $80.86_{\pm 0.12}$ &  \good{ + $1.88_{\pm 0.04}$} &  \good{ + $2.33_{\pm 0.08}$} &  \good{ + $\best{2.60}_{\pm 0.09}$} \\
& ViT-B-16 & $92.77_{\pm 0.07}$ &  \bad{$-0.12_{\pm 0.05}$} &  \good{ + $\best{0.19}_{\pm 0.09}$} &  \good{ + $\best{0.13}_{\pm 0.07}$}\\
& Mixer-B-16 & $83.77_{\pm 0.08}$ &  \good{ + $0.45_{\pm 0.06}$} &  \good{ + $0.52_{\pm 0.15}$} &  \good{ + $\best{0.97}_{\pm 0.12}$} \\
\midrule
\multirow{5}{*}{\shortstack[l]{SSL:\\ CIFAR10}} & MoCo & $\best{89.25_{\pm 0.07}}$ &  \bad{$-0.03_{\pm 0.10}$} &  \bad{$-0.25_{\pm 0.06}$} &  \bad{$-0.17_{\pm 0.10}$}\\
& SimCLR &  $\best{88.66}_{\pm 0.08}$ &  \bad{$-0.05_{\pm 0.06}$} &  \good{ + $\best{0.05}_{\pm 0.04}$} &  \bad{$-0.13_{\pm 0.06}$} \\
& SimSiam &  $\best{89.86_{\pm 0.22}}$ &  \good{ + $\best{0.12_{\pm 0.26}}$} &  \good{ + $\best{0.07_{\pm 0.10}}$} &  \good{ + $\best{0.11_{\pm 0.10}}$}\\
& BarlowTwins & $\best{86.34_{\pm 0.24}}$ &  \bad{$-0.09_{\pm 0.19}$} &  \good{ + $\best{0.09_{\pm 0.15}}$} &  \good{ + $\best{0.14_{\pm 0.05}}$} \\
& BYOL &$90.32_{\pm 0.14}$ &  \good{ + $\best{0.70}_{\pm 0.05}$} &  \good{ + $0.14_{\pm 0.03}$} &  \good{ + $0.21_{\pm 0.07}$} \\
& SwaV & $87.28_{\pm 0.05}$ &  \good{ + $\best{0.09_{\pm 0.06}}$} &  \good{ + $\best{0.07_{\pm 0.12}}$} &  \good{ + $0.02_{\pm 0.06}$} \\
\midrule
\multirow{5}{*}{\shortstack[l]{SSL:\\ ImageNette}} & MoCo &  $81.74_{\pm 0.18}$ &  \good{ + $0.97_{\pm 0.10}$} &  \good{ + $0.91_{\pm 0.32}$} &  \good{ + $\best{1.40_{\pm 0.10}}$} \\
& SimCLR & $83.28_{\pm 0.22}$ &  \good{ + $\best{0.95_{\pm 0.25}}$} &  \good{ + $0.18_{\pm 0.24}$} &  \good{ + $\best{1.07_{\pm 0.13}}$} \\
& SimSiam &  $81.77_{\pm 0.14}$ &  \good{ + $\best{0.20_{\pm 0.37}}$} &  \good{ + $\best{0.33_{\pm 0.28}}$} &  \good{ + $\best{0.18_{\pm 0.26}}$} \\
& BarlowTwins &  $77.49_{\pm 0.36}$ &  \good{ + $0.20_{\pm 0.16}$} &  \good{ + $\best{0.47_{\pm 0.27}}$} &  \good{ + $\best{0.66_{\pm 0.57}}$}  \\
& BYOL & $84.16_{\pm 0.14}$ &  \good{ + $\best{0.76_{\pm 0.08}}$} &  \good{ + $0.15_{\pm 0.25}$} &  \good{ + $0.31_{\pm 0.19}$} \\
& SwaV  & $88.16_{\pm 0.31}$ &  \good{ + $\best{1.04_{\pm 0.27}}$} &  \good{ + $0.03_{\pm 0.10}$} &  \good{ + $\best{1.03_{\pm 0.09}}$}\\
\bottomrule
\end{tabular}}
\label{tab:cv_results}
\end{table}

 \subsection{Computer Vision}

\paragraph{Supervised Classification (SC).}
We evaluate the CNN architectures WideResNets \cite{wrn} with 28 layers and width 10, and PyramidNet (PN) with 110 layers and widening factor 272 \cite{pyramidnet} as well as Vision Transformer (ViT) \cite{vit} and MLP-Mixer \cite{mlp-mixer} on CIFAR\{10, 100\} \cite{cifar}. All experiments use basic data augmentations: horizontal flip, padding by four pixels, random crop, and cutout \cite{cutout}. In \Cref{sec:da}, we experiment with different data augmentation schemes. 

\paragraph{Self-Supervised Learning (SSL).}
We consider the following methods on CIFAR10 and ImageNette\footnote{\url{https://github.com/fastai/imagenette}}: Momentum Contrast \cite{moco}, a Simple framework for Contrastive Learning (SimCLR) \cite{simclr}, Simple Siamese representation learning (SimSiam) \cite{simsiam}, Barlow Twins \cite{barlow_twins}, Bootstrap your own Latent (BYOL) \cite{byol}, and Swapping Assignments between multiple Views of the same image (SwAV) \cite{swav}. All SSL methods use a ResNet-18 \cite{resnet} as backbone network. To test the frozen representations, we use $k$-nearest-neighbor classification with a memory bank \cite{knn_ssl}. We choose $k\!=\!200$ and temperature $\tau\!=\!0.1$ to reweight similarities. Compared to learning a linear model on top of the representations, this evaluation procedure is more robust to hyperparameter changes \cite{revisiting_ssl}.

\subsection{Natural Language Processing}
We consider the task of open domain question answering (ODQA) using a T5-based  model Fusion-In-Decoder (FiD) \cite{izacard2021leveraging}. 
We evaluate FiD-base on the test sets of Natural Questions (NQ)~\cite{kwiatkowski2019natural} and TriviaQA \cite{joshi2017triviaqa}. We also consider natural language understanding tasks included in the GLUE benchmark \cite{wang2018glue}, which cover acceptability, sentiment, paraphrase, similarity, and inference. We fine-tune RoBERTa-base \cite{liu2019roberta} for each task individually and report the results on the GLUE dev set.

\begin{figure}
    \centering
        \caption{(a) NLP test results: Open-Domain Question Answering and Natural Language Understanding (GLUE) including paraphrase, sentiment analysis, and textual entailment. (b) GRL test results: Node Property Prediction (NPP), Graph Property Prediction (GPP), Link Property Prediction (LPP).}
    \begin{subfigure}{0.49\textwidth}
    \scalebox{0.55}{
\begin{tabular}{llcccccccc} \toprule
Task & Model & \multicolumn{1}{c}{ Baseline } & SWA & SAM &  WASAM\\ 
\midrule
\multirow{1}{*}{NQ} & FiD & $49.35_{\pm 0.44}$ & \bad{$- 0.20_{\pm 0.33}$}&
\good{+ $\best{0.33_{\pm 0.19}}$} & \good{+ $\best{0.48_{\pm 0.21}}$ }& \\
\midrule
\multirow{1}{*}{TriviaQA} & FiD&
$67.74_{\pm 0.29}$& \good{+ $0.40_{\pm 0.24}$}&
\good{+ $\best{0.89_{\pm 0.03}}$} & \good{+ $\best{0.92_{\pm 0.10}}$} &  \\
\midrule
COLA & RoBERTa &
$60.41_{\pm 0.22}$& \good{+ $0.09_{\pm 0.08}$}&
\good{+ $\best{1.57_{\pm 1.20}}$} &\good{+ $\best{1.41_{\pm 1.14}}$}& \\
\midrule
SST & RoBERTa &
$94.95_{\pm 0.13}$& \bad{$- 0.30_{\pm 0.27}$}&
\bad{$- 0.23_{\pm 0.40}$}&\good{+ $\best{0.19_{\pm 0.14}}$}& \\
\midrule
MRPC & RoBERTa &
$89.14_{\pm 0.57}$& \good{+ $0.08_{\pm 0.49}$}&
\good{+$\best{0.73_{\pm 0.43}}$} & \good{+ $\best{\mathbf{0.81}_{\pm 0.38}}$}& \\
\midrule
STSB & RoBERTa &
$90.40_{\pm 0.02}$& \good{+$0.00_{\pm 0.05}$}&
\good{+$\mathbf{0.38}_{\pm 0.17}$}& \good{+ $\best{0.35_{\pm 0.16}}$}& \\
\midrule
QQP & RoBERTa &
$91.36_{\pm 0.07}$&\good{+$0.01_{\pm 0.06}$}&
\good{+$\best{0.08_{\pm 0.07}}$}&\good{+$\best{0.06_{\pm 0.08}}$}& \\
\midrule
MNLI & RoBERTa &
$87.41_{\pm 0.09}$&\good{+$0.08_{\pm 0.11}$}&
\good{+$\best{0.39_{\pm 0.02}}$}&\good{+$0.35_{\pm 0.03}$}& \\
\midrule
QNLI & RoBERTa &
$92.96_{\pm 0.06}$&\bad{$- 0.08_{\pm 0.11}$}&
\good{+$0.09_{\pm 0.01}$}& \good{+$\best{0.11_{\pm 0.06}}$}& \\
\midrule
RTE & RoBERTa &
$80.09 _{\pm 0.23}$&\bad{$- 0.23_{\pm 0.20}$}&
\good{+$\best{0.70_{\pm 0.65}}$}& \bad{$- 0.46_{\pm 0.12}$}& \\
\bottomrule
\end{tabular}
}
    \caption{}
    \label{tab:nlp_results}
    \end{subfigure}
    \begin{subfigure}{0.49\textwidth}
    \scalebox{0.55}{
\begin{tabular}{llcccccccc} \toprule
Task & Model & \multicolumn{1}{c}{ Baseline } & SWA & SAM &  WASAM\\ 
\midrule
\multirow{2}{*}{\shortstack[l]{NPP:\\ Proteins}} & SAGE & $\best{77.79_{\pm 0.18}}$ &  \bad{$\best{-0.17_{\pm 0.22}}$} &  \bad{$\best{-0.02_{\pm 0.13}}$} & \bad{$\best{-0.11_{\pm 0.15}}$}\\
& DGCN & $\best{85.42_{\pm 0.17}}$ &  \good{+ $\best{0.11_{\pm 0.08}}$} &  \bad{$-0.14_{\pm 0.05}$} &  \bad{$-0.08_{\pm 0.07}$} \\
\midrule
\multirow{2}{*}{\shortstack[l]{NPP:\\ Products}} & SAGE &$78.92_{\pm 0.08}$ &  \good{ + $0.39_{\pm 0.10}$} &  \good{ + $0.13_{\pm 0.08}$} &  \good{ + $\best{0.57_{\pm 0.03}}$} \\
& DGCN & $73.88_{\pm 0.13}$ &  \good{ + $\best{0.44_{\pm 0.14}}$} &  \good{ + $0.08_{\pm 0.09}$} &  \good{ + $\best{0.53_{\pm 0.05}}$}\\
\midrule
\multirow{2}{*}{\shortstack[l]{GPP:\\ Code2}} & GCN & $16.04_{\pm 0.09}$ &  \good{ + $0.73_{\pm 0.11}$} &  \good{ + $0.36_{\pm 0.08}$} &  \good{ + $\best{0.93_{\pm 0.15}}$} \\
 & GIN & $15.73_{\pm 0.11}$ &  \good{ + $0.83_{\pm 0.11}$} &  \good{ + $0.57_{\pm 0.09}$} &  \good{ + $\best{1.10_{\pm 0.09}}$} \\
 \midrule
\multirow{2}{*}{\shortstack[l]{GPP:\\ Molpcba}} & GIN & $28.10_{\pm 0.11}$ &  \good{ + $\best{0.40_{\pm 0.18}}$} &  \bad{$-0.33_{\pm 0.14}$} &  \good{ + $\best{0.33_{\pm 0.16}}$} \\
 & DGCN & $25.65_{\pm 0.13}$ &  \good{ + $\best{1.90_{\pm 0.20}}$} &  \bad{$-0.13_{\pm 0.18}$} &  \good{ + $1.34_{\pm 0.12}$}  \\
\midrule
\multirow{2}{*}{\shortstack[l]{LPP:\\ Biokg}} & CP & $84.06_{\pm 0.00}$ &  \good{ + $\best{0.07_{\pm 0.01}}$} &  \good{ $0.00_{\pm 0.03}$} &  \good{ + $\best{0.08_{\pm 0.02}}$}  \\
 & ComplEx & $84.94_{\pm 0.01}$ &  \good{ + $\best{0.14_{\pm 0.01}}$} &  \bad{$-0.02_{\pm 0.01}$} &  \good{ + $\best{0.12_{\pm 0.02}}$} \\
\midrule
\multirow{2}{*}{\shortstack[l]{LPP:\\ Citation2}} & GCN & $79.52_{\pm 0.41}$ &  \bad{$-0.05_{\pm 0.52}$} &  \good{ + $1.32_{\pm 0.06}$} &  \good{ + $\best{1.50_{\pm 0.13}}$} \\
 & SAGE &  $81.95_{\pm 0.02}$ &  \good{ + $\best{1.15_{\pm 0.02}}$} &  \bad{$-0.31_{\pm 0.07}$} &  \good{ + $0.86_{\pm 0.04}$} \\
\bottomrule
\end{tabular}
}
    \caption{
    }
    \label{tab:grl_results}
    \end{subfigure}
    \label{fig:my_label}
\end{figure}

\subsection{Graph Representation Learning}
We use a subset of the Open Graph Benchmark (OGB) datasets \cite{ogb}. The tasks are node property prediction (NPP), graph property prediction (GPP), and link property prediction (LPP). For each task, we use two of the following GNN architectures and matrix factorization methods: GCN \cite{GCN}, DeeperGCN (DGCN) \cite{deepergcn}, SAGE \cite{SAGE}, GIN \cite{GIN}, ComplEx \cite{Complex}, and CP \cite{CP}.
We use popular training schemes, such as virtual nodes, cluster sampling \cite{clustergcn}, or relation prediction as auxiliary training objective \cite{rp}. 
The reported metrics are ROC-AUC for Proteins, Accuracy for Products, F1 score for Code2, Average precision for Molpcba, and Mean Reciprocal Rank for Biokg/Citation2.

\subsection{9 Findings}
We use $\gG(\cdot)$ to describe the generalization accuracy/F1/ROCAUC/AP/MRR of all optimizers $\{\text{Non-flat baseline} (\NF),\SWA,\SAM,\SWASAM\}$.

\begin{enumerate}[leftmargin = *] 

\item \textbf{Datasets matter.} For example, for node property prediction (Proteins), we see that no flat optimizer improves over the baseline optimizer; however, for (Products), flat-minima optimizers on the same architectures significantly improve over the baseline. We further explore the impact of different data augmentation strategies in \Cref{sec:da}.

\item \textbf{Architectures matter,} e.g., there is a vast difference across model architectures for link property prediction on the Citation2 dataset: using a GNN with GCN layers, SAM achieves a statistically significant boost of >$1.30\%$ and SWA slightly hurts the performance. When we replace the GCN layers with SAGE layers (and fix everything else), we see a boost of $>1.15\%$ for SWA, while SAM hurts the performance by $-0.31\%$. 
\item \textbf{SWA underperforms on NLP tasks.} SAM achieves the best performance in 7/10 experiments on NLP tasks, consistent with the findings of \citet{SAM_NLP}, which show that SAM can boost performance across a wide range of NLP tasks. However, SWA never performs best, only improving the results in 1/10 cases, and even hurting the performance on 4 tasks. Surprisingly, $\gG(\SWASAM) > \gG(\SAM)$ for \{SST, QNLI\} while SWA decreases performance in these cases. 
\item \textbf{SWA beats SAM on GRL tasks.} $\gG(\SWA) > \gG(\NF)$ in 10/12 experiments, while $\gG(\SAM) > \gG(\NF)$ only in 4. We examine why SAM under-performs in \Cref{app:sam_grl}.  
\item \textbf{SWA does not work well with Transformers.} SWA often does not improve and sometimes hurts performances, as can be seen in the ViT results in \Cref{tab:cv_results}, and NLP results in \Cref{tab:nlp_results}. In contrast, SAM has some positive effects in these settings. 
We explore this further in \Cref{app:swa_nlp}.
\item \textbf{SWA and SAM improve SSL task performance.} This is non-trivial as the theoretical motivation behind finding flat minima is linked to supervised learning losses \cite{hochreiter1997flat,relative_flatness,PACBayes}. Concurrently to our work, \citet{ramesh2022hierarchical} report that SAM helps for contrastive CLIP \cite{CLIP} models too.
\item \textbf{Flat optimizers do not strictly improve over non-flat optimizers.} The non-flat optimizer is nearly always the best for  NPP Proteins and SSL methods on CIFAR10. We investigate the NPP Proteins solutions in the next subsection and recommend a more thorough investigation of the landscapes of SSL objectives for future work.
\item \textbf{Flat-minima optimizers offer asymmetric payoffs:} at worst, they decreased performance by $-0.30\%$, at best, they increased it by $2.60\%$. 
\item \textbf{Averaging SAM iterates often improves over SWA or SAM alone}. 
$\gG(\SWASAM) > \min(\gG(\SWA), \gG(\SAM))$ in 39/42 cases. We hypothesize that asymmetric payoffs are the reason: when either SWA or SAM does not improve over the baseline (as discussed above), it does not hurt (much) either, hence WASAM is more robust across all tasks.
\end{enumerate}

\subsection{Why do flat-minima optimizers fail?}\label{subsec:fail} 
Here, we audit one of the cases, where neither $\swasol$ nor $\samsol$ improves over $\nfsol$ (this happens in 3 out of 42 cases): training a GraphSAGE \cite{SAGE} model on OGB-Proteins: a protein-protein interaction graph where the goal is to predict the presence of protein functions (multi-label binary classification) \cite{ogb}. $\swasol$ performs noticeably worse; $\samsol$ performs about equally well.

\begin{figure*}[h!]
    \centering
    \includegraphics[width=\textwidth]{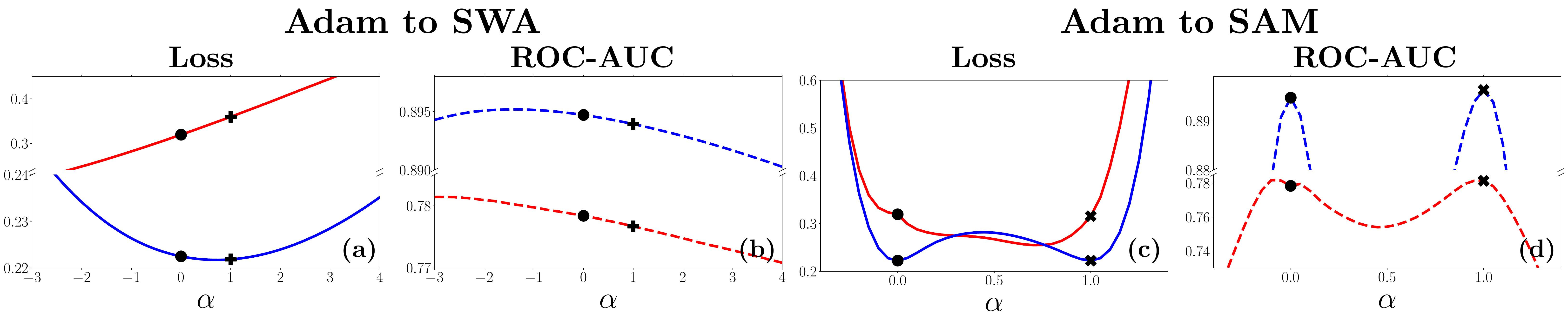}
    \caption{GraphSAGE on OGB-Proteins: Adam's ($\bullet$) solution performs about equally well as SAM ($\bm{\times}$), and better than SWA ($\bm{+}$).}
    \label{fig:failure}
\end{figure*}
\Cref{fig:failure} shows two linear interpolations: between $\nfsol$ (ADAM) and (1) $\swasol$ (Figures~\ref{fig:failure}a and \ref{fig:failure}b), and (2) $\samsol$ (Figures~\ref{fig:failure}c and \ref{fig:failure}d). In contrast to success cases in \Cref{fig:flat_vs_non_flat}, here: (a) for both SWA and SAM, the training loss minimizer is very uncorrelated with the test loss minimizer; (b) SAM and ADAM seem to be contained in the same test loss/accuracy basin. More analyses can be found in the Appendix.

\section{Limitations and Future Work} \label{sec:discussion}
First, some of the fixed, shared hyperparameter values we used from previous works may harm the effect of flat optimizers. The ideal experimental design includes tuning all hyperparameters independently for the non-flat optimizer, SWA, SAM, and WASAM. However, this forces the number of required runs to grow exponentially in unique hyperparameters and quickly renders this benchmark infeasible. 

Second, despite our best efforts to evaluate the optimizers on a broad range of benchmark tasks, there are still plenty of unexplored domains; especially some of which are known to be sensitive to careful optimization. For example, bi-level optimization problems \cite{franceschi2018bilevel} are common in generative modeling \cite{goodfellow2020generative,GANs_nash}, deep reinforcement learning \cite{konda1999actor,hong2020two}, meta-learning \cite{rajeswaran2019meta,NEURIPS2020_ef0d17b3}, or causal machine learning \cite{kaddour2021causal,survey}. We are unaware of an investigation of flat minima optimization for such problems. 

Third, in general, we believe fruitful directions of research include (a) optimizers that explicitly find basins where training loss flatness more directly corresponds to higher hold-out accuracy, (b) post-processing methods for existing optimization runs to move into flatter regions of these basins \cite{andriushchenko2022towards}, (c) loss functions whose contours more tightly align with accuracy contours, (d) the study of flat-minima hyperparameter interactions (e.g., learning rate and neighborhood radius in SAM) (see \Cref{app:sam_hparam,app:base_optim_hparam} for first results), (e) analyses of flat minima optimization on convergence speed \cite{lawa}.

Our benchmark results point to which tasks would most benefit from improving these future work directions: graph learning tasks would benefit from improvements in (a), as SAM is never among the best-performing method, and language tasks would benefit if (b) is improved, as SWA is never among the best performing method).

\section{Conclusion}
We investigated when flat minima optimizers work by conducting a fair comparison of two popular flat-minima optimizers. We examined the behavior of SWA/SAM by analyzing their loss landscapes on two representative deep learning tasks. Our next step was to evaluate their generalization performance on a broad and diverse set of tasks (in data, learning settings, and model architectures). Based on this benchmarking, we identified 9 findings, of which some directly guide future work directions. Finally, when SWA/SAM did not improve over baselines, common assumptions seemed broken (i.e., train-to-test loss minimizers were not correlated).

\section*{Acknowledgements}
We are very grateful to Kilian Q. Weinberger and Gao Huang for initial discussions and intuition, Pontus Stenetorp for NLP experimental design advice, and Oscar Key for feedback on the draft. JK and LL acknowledge support by the Engineering and Physical Sciences Research Council with grant number EP/S021566/1. This research was supported through Azure resources provided by The Alan Turing Institute and credits awarded by Google Cloud.

\bibliographystyle{icml2021}
\bibliography{references}

\section*{Checklist}


\begin{enumerate}

\item For all authors...
\begin{enumerate}
  \item Do the main claims made in the abstract and introduction accurately reflect the paper's contributions and scope?
    \answerYes{}
  \item Did you describe the limitations of your work?
    \answerYes{See \Cref{sec:discussion}}
  \item Did you discuss any potential negative societal impacts of your work?
    \answerNo{We consider this work to be an investigation into optimization algorithms that are central to modern ML systems. As these systems can be used in vastly different ways, we believe that it is intractable to isolate the societal impact of novel insights into optimization.}
  \item Have you read the ethics review guidelines and ensured that your paper conforms to them?
    \answerYes{}
\end{enumerate}

\item If you are including theoretical results...
\begin{enumerate}
  \item Did you state the full set of assumptions of all theoretical results?
    \answerNA{}
        \item Did you include complete proofs of all theoretical results?
    \answerNA{}
\end{enumerate}

\item If you ran experiments...
\begin{enumerate}
  \item Did you include the code, data, and instructions needed to reproduce the main experimental results (either in the supplemental material or as a URL)?
    \answerYes{Supplemental material}
  \item Did you specify all the training details (e.g., data splits, hyperparameters, how they were chosen)?
    \answerYes{See Appendix.}
        \item Did you report error bars (e.g., with respect to the random seed after running experiments multiple times)?
    \answerYes{See \Cref{sec:benchmark}}
        \item Did you include the total amount of compute and the type of resources used (e.g., type of GPUs, internal cluster, or cloud provider)?
    \answerYes{See Appendix.}
\end{enumerate}

\item If you are using existing assets (e.g., code, data, models) or curating/releasing new assets...
\begin{enumerate}
  \item If your work uses existing assets, did you cite the creators?
    \answerYes{}
  \item Did you mention the license of the assets?
    \answerYes{See Appendix.}
  \item Did you include any new assets either in the supplemental material or as a URL?
    \answerNA{}
  \item Did you discuss whether and how consent was obtained from people whose data you're using/curating?
    \answerNA{}
  \item Did you discuss whether the data you are using/curating contains personally identifiable information or offensive content?
    \answerNA{}
\end{enumerate}

\item If you used crowdsourcing or conducted research with human subjects...
\begin{enumerate}
  \item Did you include the full text of instructions given to participants and screenshots, if applicable?
    \answerNA{}
  \item Did you describe any potential participant risks, with links to Institutional Review Board (IRB) approvals, if applicable?
    \answerNA{}
  \item Did you include the estimated hourly wage paid to participants and the total amount spent on participant compensation?
    \answerNA{}
\end{enumerate}

\end{enumerate}


\appendix
\newpage

\appendix

\section{Additional Analyses}

\subsection{CKA Similarities} \label{app:cka}

We compute the CKA \cite{cka_similarity, taxonomy} and cosine similarities of network output logits on train and test set, respectively. \Cref{tab:cka_similarities} shows the results. 

\begin{table}[h]
\centering
    \caption{\textbf{Pairwise CKA \cite{cka_similarity} and cosine similarities} between non-flat (NF) and SWA/SAM solutions. SWA solutions produce predictions more similar to NF ones than SAM.  }

\resizebox{\columnwidth}{!}{%

\begin{tabular}{l|c|c|c|c}
         \toprule
         \bf Task & $\cka(\nfsol, \swasol)$ & $\cossim(\nfsol, \swasol)$ & $\cka(\nfsol, \samsol)$ & $\cossim(\nfsol, \samsol)$ \\ \midrule  
         WRN-CIFAR100 (Train) & 0.9880 & 0.9812 & 0.9810 & 0.9240 \\
         WRN-CIFAR100 (Test) & 0.9137 & 0.9732 & 0.8580 & 0.9045 \\
         \midrule
         GIN-Code2 (Train) & 0.8522 & 0.9730 & 0.7276 & 0.9515 \\
         GIN-Code2 (Test) & 0.8677 & 0.9750 & 0.7275 & 0.9516\\
         \midrule
         RoBERTa-QNLI (Train) & 0.9997 &  0.9991 & 0.9790  & 0.9510 \\
         RoBERTa-QNLI (Valid) & 0.9830 & 0.9959 & 0.9550 & 0.9530\\
         \midrule
        RoBERTa-RTE (Train) & 0.9931 & 0.9891 & 0.9831 & 0.9628\\
         RoBERTa-RTE (Test) & 0.9314 & 0.9567 & 0.8808 &  0.8927\\
         \midrule
         GIN-Molpcba (Train) & 0.8886 & 0.9973 & 0.7441 & 0.9804 \\
         GIN-Molpcba (Test) & 0.8772 & 0.9942 & 0.7232 & 0.9730 \\
        \bottomrule
    \end{tabular}}
    \label{tab:cka_similarities}
\end{table}

The results show that the SAM solutions produce predictions that are less similar to the non-flat baseline than SWA solutions, as indicated by lower CKA and cosine similarities. This result is in line with Observation 1 and 2 from \Cref{subsec:interpolations}. 

\subsection{Saddle point of SAM's GIN solution} \label{app:sam_saddle}

\begin{figure}[h]
    \centering
    \begin{subfigure}[t]{0.49\textwidth}
        \includegraphics[width=\textwidth]{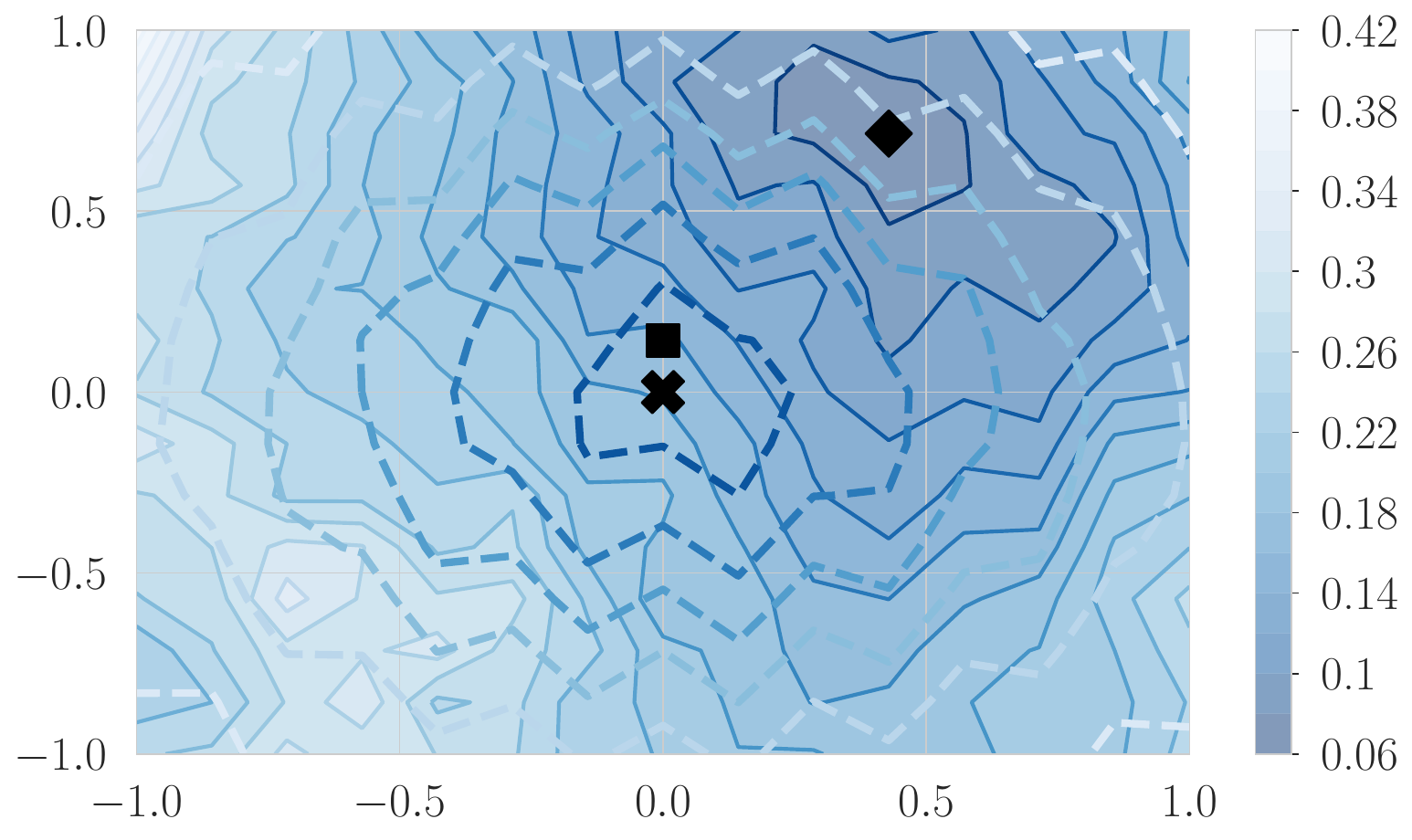}
        \caption{SAM solution.}
    \end{subfigure}
    \begin{subfigure}[t]{0.49\textwidth}
        \includegraphics[width=\textwidth]{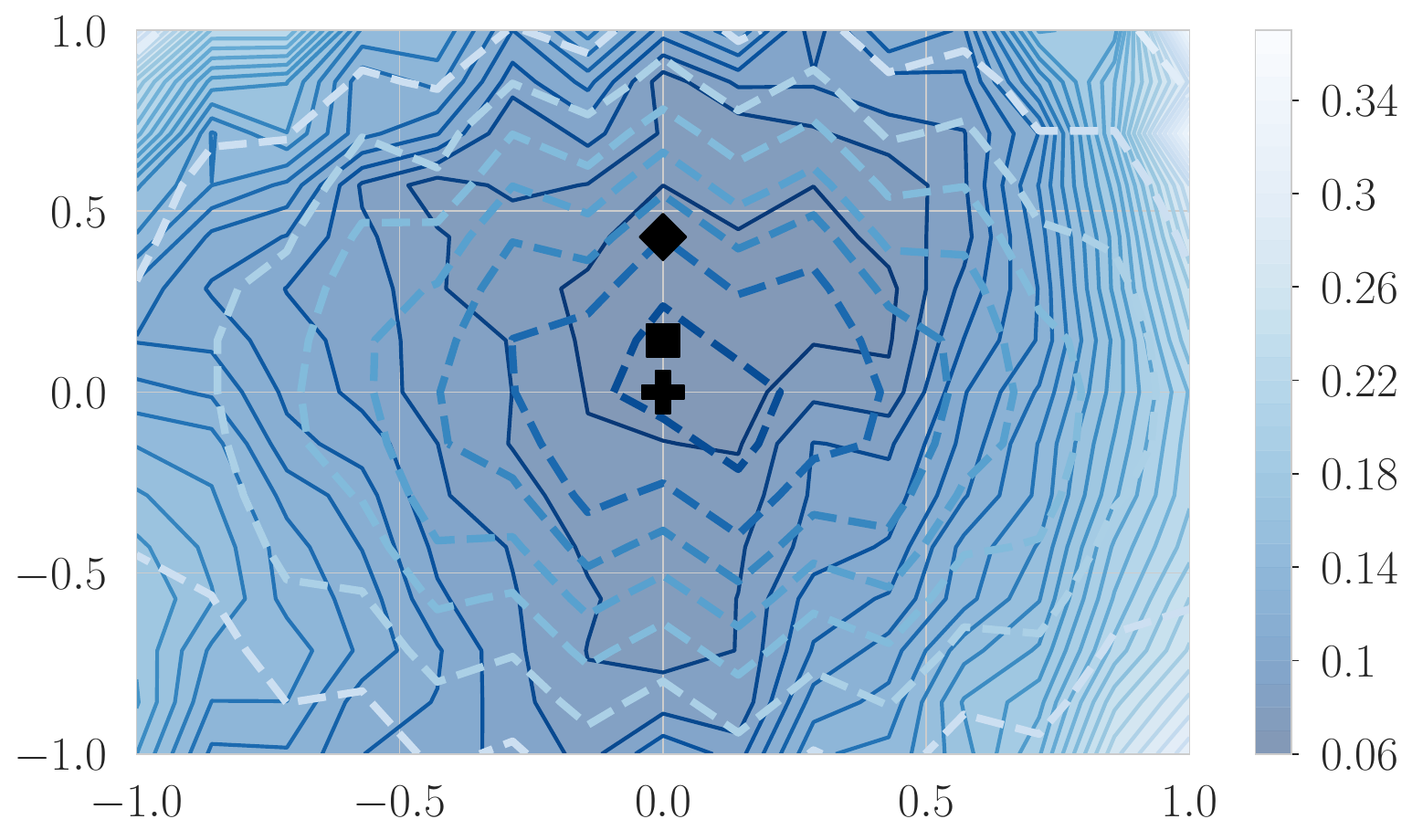}
        \caption{SWA solution (for comparison).}
    \end{subfigure}
    \caption{\textbf{SAM and SWA solution in 2D random plane for GIN-Code2 task.} We depict $\samsol, \swasol$ by $\bm{\times}, \bm{+}$, respectively, the test set F1 score maximizer  by $\blacksquare$, and the training loss minimizer by $\blacklozenge$. The converged SAM solution is distant from the training loss minimizer in the 2D plane: $\trainloss(\samsol) = 0.1779 \gg 0.0672 = \trainloss(\params^{\blacklozenge})$. Further, losses (---) and F1 scores (\protect\makebox[1.3em]{\protect\xdotfill{.4pt}}) are not well-aligned. In contrast, the SWA solution is almost the training loss minimizer: $\trainloss(\swasol) = 0.0661 \approx 0.0609 = \trainloss(\params^{\blacklozenge})$. Also, losses and F1 scores are better aligned. Yet, $\samsol$ and $\swasol$ perform about equally well on the test set, see \Cref{fig:wasam_works}d.}
    \label{fig:sam_saddle}
\end{figure}

\begin{figure}[h]
    \centering
    \includegraphics[width=0.8\textwidth]{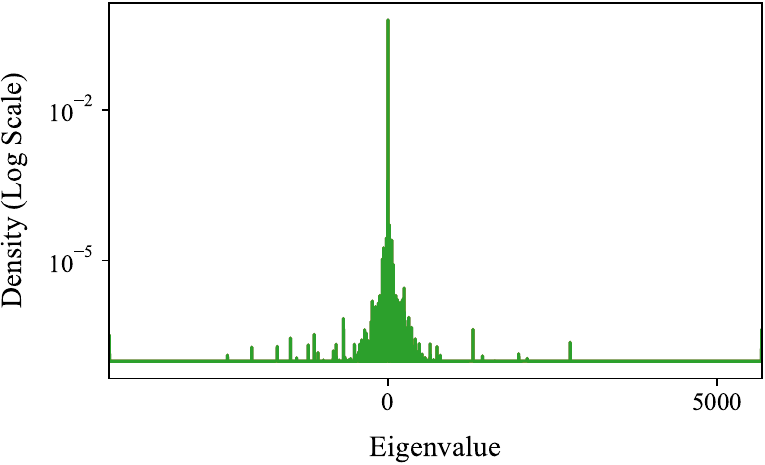}
    \caption{\textbf{Hessian Eigenvalue Density \cite{eigenvalue_density} of SAM's GIN-Code2 solution}: We observe significant probability mass around both positive and negative eigenvalues, indicating that this solution is a saddle point.}
    \label{fig:esd}
\end{figure}

To gather further evidence on whether the SAM solution is a saddle point, we analyze its Hessian eigenvalue density, following \citet{eigenvalue_density} and using the Stochastic Lanczos Quadrature algorithm \cite{golub1969calculation}. \Cref{fig:esd} shows the density, including significant probability mass for both positive and negative eigenvalues, indicating that the solution is a saddle point.

\subsection{Why does SWA not work well on NLP tasks?} \label{app:swa_nlp}
In \Cref{tab:nlp_results}, we saw that SWA had only a mild effect on the generalization performance of NLP tasks, sometimes even decreasing it. Here, we seek to investigate why that is. 

We consider two tasks: (i) the RTE task, for which SWA decreases the performance by around $-0.23_{\pm 0.20}$ compared to Adam, (ii) the QNLI task, for which SWA decreases the performance by $-0.08_{\pm 0.11}$. In both cases, SAM improved the performance statistically significantly over Adam.  
\begin{figure}[h]
    \centering
    \begin{subfigure}[b]{\textwidth}
    \caption{NLP: RoBERTa-QNLI}
    \includegraphics[width=\textwidth]{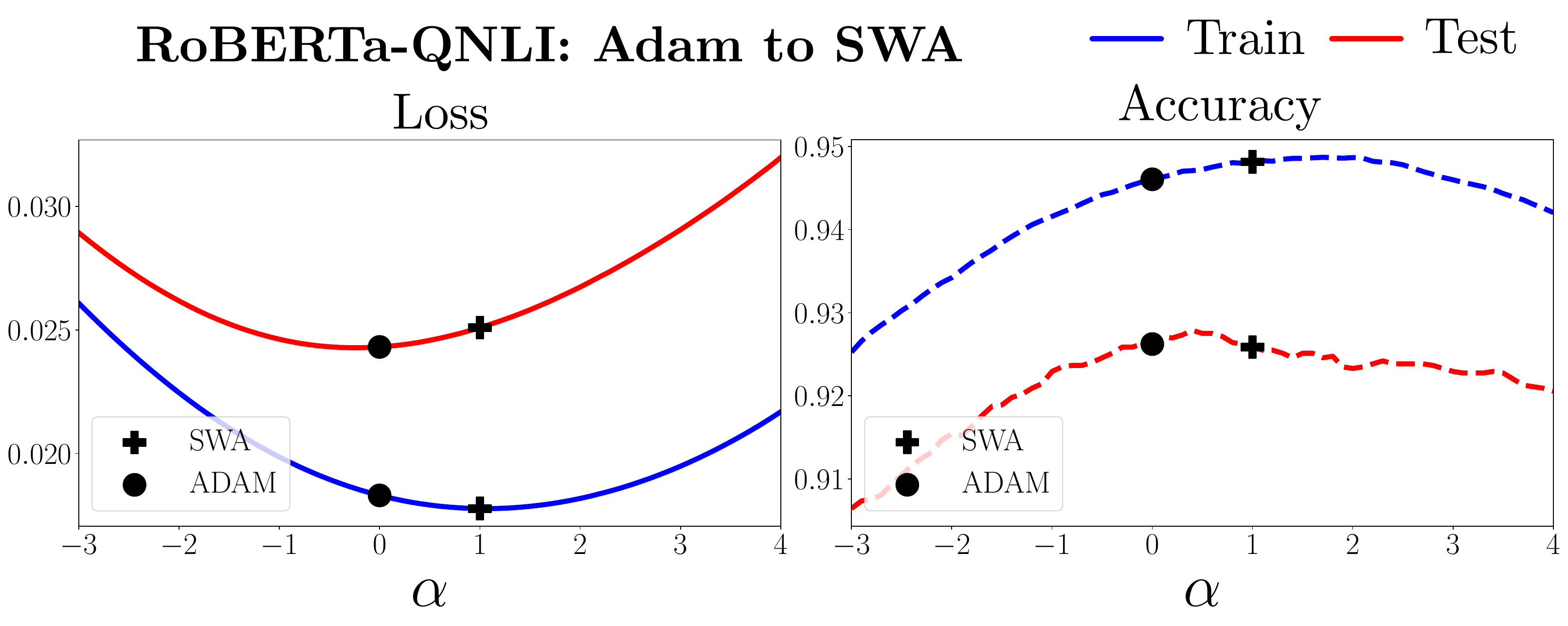}
    \label{failure:qnli}
    \end{subfigure}
    \begin{subfigure}[b]{\textwidth}   
    \caption{NLP: RoBERTa-RTE}
    \includegraphics[width=\textwidth]{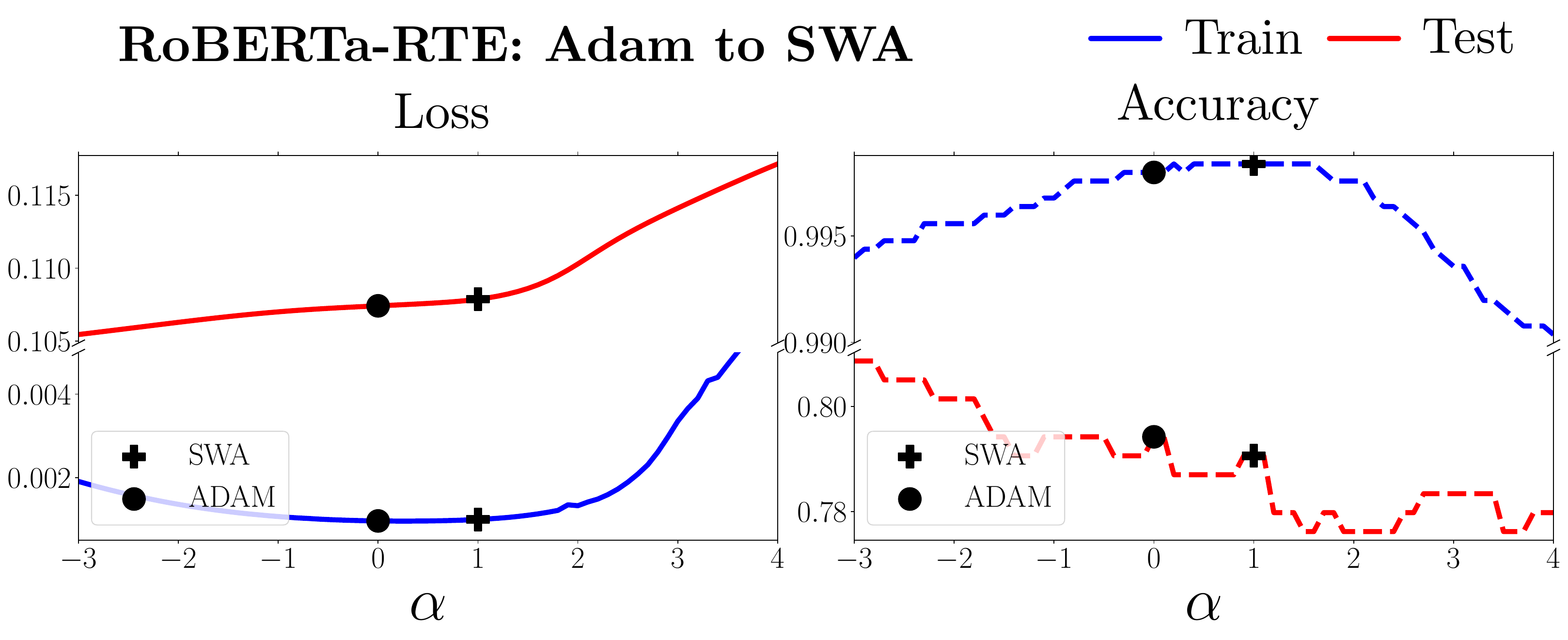}
    \label{failure:rte}
    \end{subfigure}
	\caption{Training (\textcolor{blue}{blue}) and test (\textcolor{red}{red}) losses (---) and accuracies (\protect\makebox[1.3em]{\protect\xdotfill{.4pt}}) of linear interpolations $\params(\alpha)=(1-\alpha) \params+\alpha \params^{\prime}$ between Adam solutions ($\bullet, \alpha=0.0$) and SWA ($\bm{+}, \alpha=1.0$).}
    \label{fig:failure_swa}
\end{figure}

For the QNLI task in \Cref{failure:qnli}, we observe that SWA finds a lower/higher training loss/accuracy than Adam, respectively. However, the test loss/accuracy is higher/lower at the SWA solutions and the loss functions seem less well correlated in between both solutions (i.e, for $\alpha \in [0,1]$).

For the RTE task in \Cref{failure:rte}, we note that SWA finds a solution that is closer to a sharply increasing side. This may happen if the baseline optimizer skips or goes around sharper solutions (e.g., due to large step sizes) and the average pulls it towards these suboptimal regions. 

Further, in \Cref{tab:cka_similarities}, we notice very high values of $\cka(\nfsol, \swasol), \cossim(\nfsol, \swasol)$ for both training and test sets, indicating that the predictions are indeed very similar. In contrast, $\cka(\nfsol, \samsol), \cossim(\nfsol, \samsol)$ is lower, especially for the test set. 

\subsection{Why does SAM not work well on GRL tasks?} \label{app:sam_grl}

\begin{figure}[h]
    \centering
    \caption{GPP: Molpcba-GIN}
    \begin{subfigure}[b]{\textwidth}
    \includegraphics[width=\textwidth]{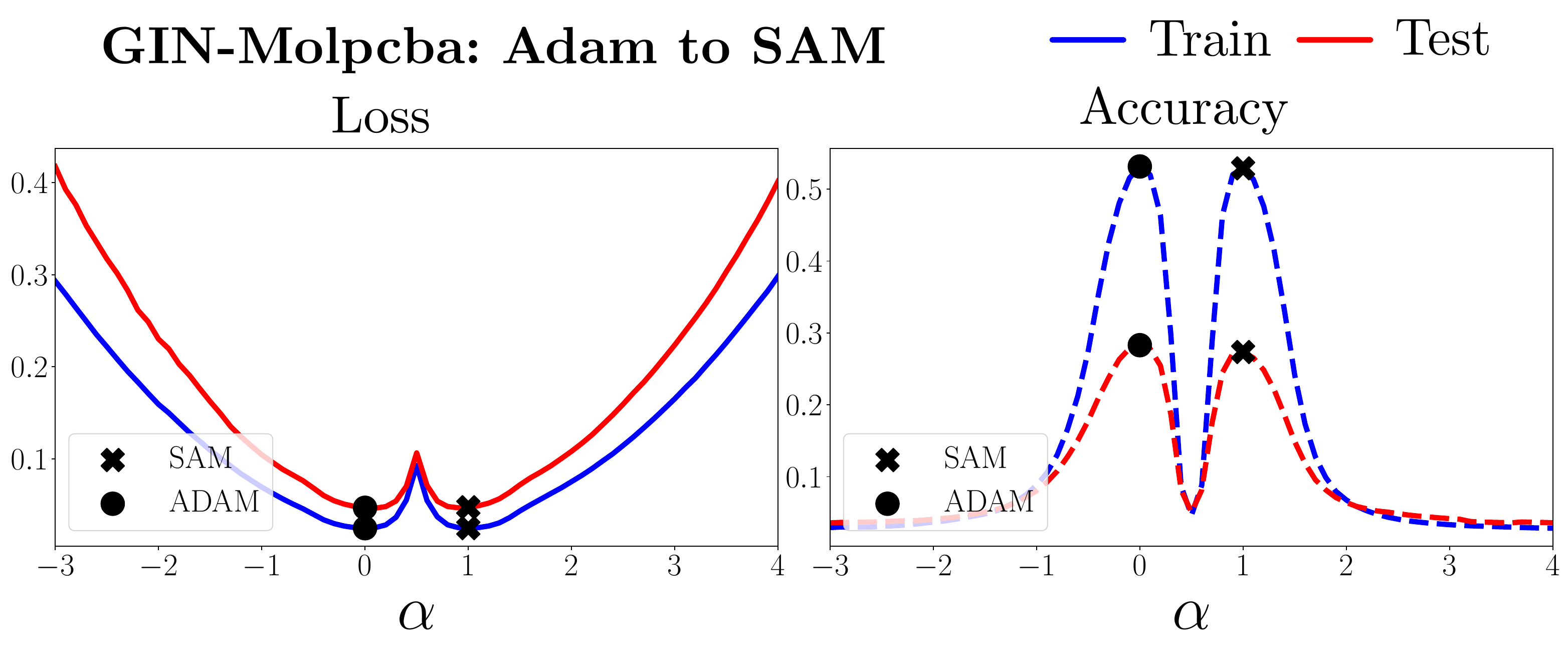}
    \end{subfigure}
	\caption{Training (\textcolor{blue}{blue}) and test (\textcolor{red}{red}) losses (---) and accuracies (\protect\makebox[1.3em]{\protect\xdotfill{.4pt}}) of linear interpolations $\params(\alpha)=(1-\alpha) \params+\alpha \params^{\prime}$ between Adam solutions ($\bullet, \alpha=0.0$) and SAM ($\bm{\times}, \alpha=1.0$).}
    \label{fig:failure_sam}
\end{figure}

\Cref{fig:failure_sam} shows the interpolations between the Adam and SAM solution. We do not observe a significant loss/accuracy difference between the two different basins. One possible explanation for this phenomenon is that the loss surface for this task is ``globally well-connected'' \cite{taxonomy}, yielding many basins with very similar geometric properties. 

\newpage
\clearpage

\subsection{Does changing SAM's $\rho$ result in different basins?} \label{app:sam_hparam}
Here, we plot linear interpolations of solutions obtained by smaller and larger $\rho$ values. Overall, we find that all solutions seem to lie in different basins indicated by high loss barriers in between ($\alpha=0.5$) them.

\subsubsection{WRN-28-10}
For the WRN-28-10 model investigated in \Cref{subsec:interpolations} and \Cref{sec:benchmark}, we set $\rho=0.1$ (as determined by hyper-parameter tuning on validation loss). 
\begin{figure}[h]
    \centering
    \caption{\textbf{WRN-28-10}: Changing SAM's $\rho$ result in different basins.}
    \includegraphics[width=\textwidth]{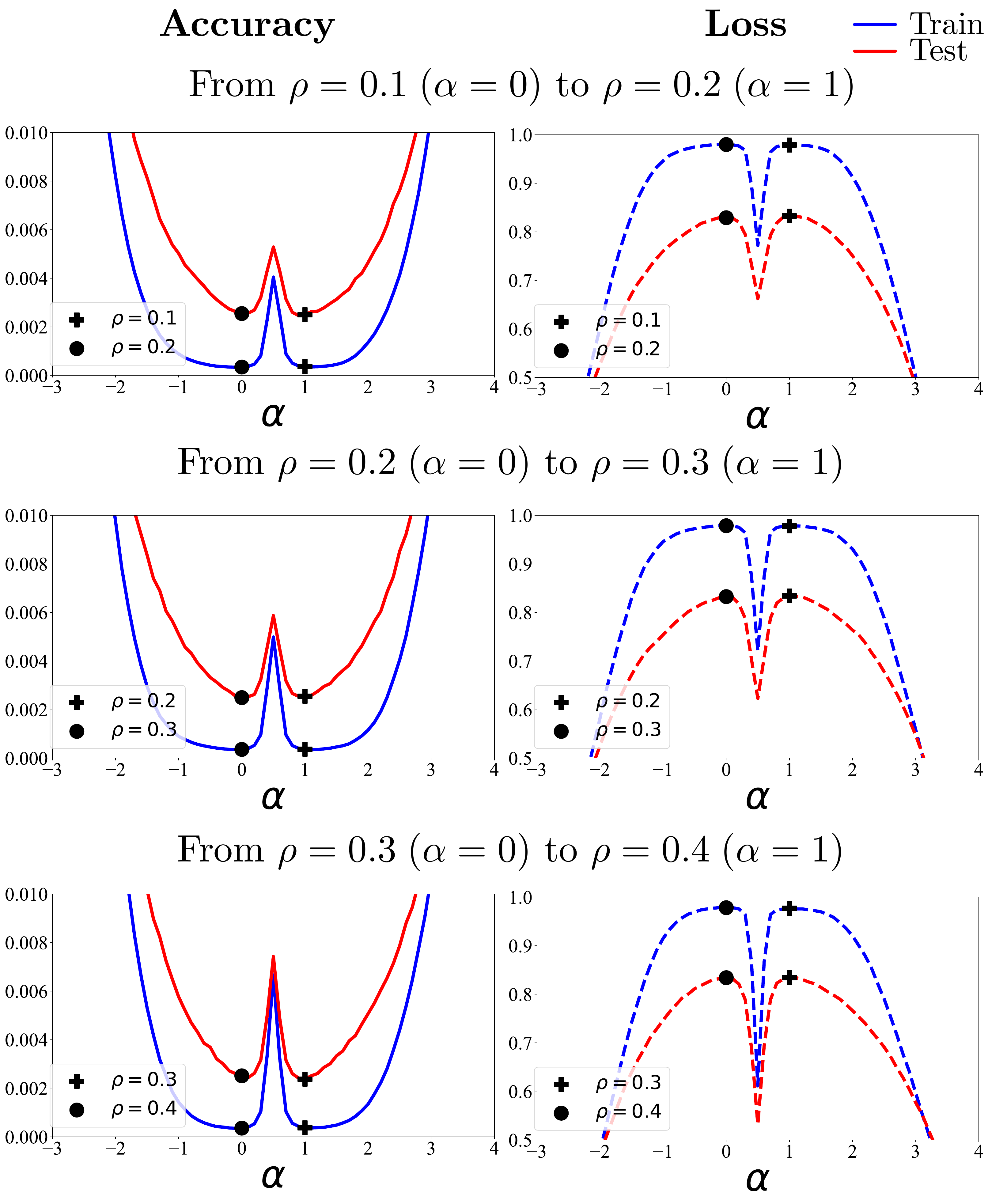}
\end{figure}

\pagebreak
\clearpage

\subsubsection{GIN-Code2}
For the GIN model investigated in \Cref{subsec:interpolations} and \Cref{sec:benchmark}, we set $\rho=0.15$ (as determined by hyper-parameter tuning on validation loss). 

\begin{figure}[h]
    \centering
    \caption{\textbf{GIN-Code2}: Changing SAM's $\rho$ result in different basins.}
    \includegraphics[width=\textwidth]{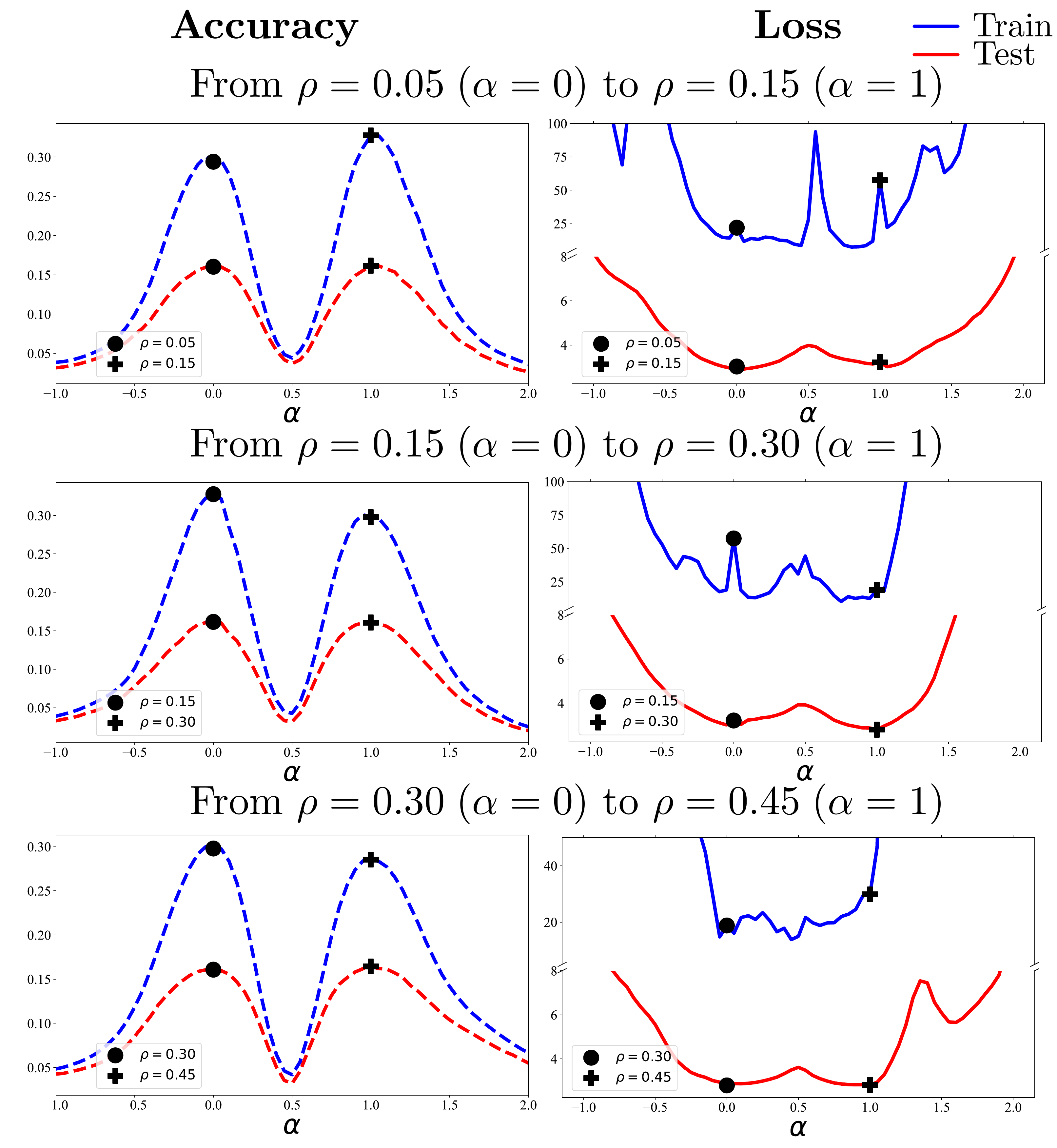}
\end{figure}

\pagebreak
\clearpage

\subsection{Does changing the base optimizer impact SWA's and SAM's effectiveness?}
\label{app:base_optim_hparam}
The influence of the base optimizer on SWA/SAM's effectiveness is under-explored. In the previous experiments, we use the default base optimizers from existing code repositories or, as reported in previous works. Here, we want to conduct an initial investigation into its effect on SWA and SAM.  

In the following experiments, we only switch the base optimizer and keep everything else fixed (including hyper-parameters such as learning rate, etc.). We train (i) a ResNet-34 (similar to WRN-28-10 but smaller) on CIFAR100, once per SGD with momentum and AdamW \cite{adamW}, and (ii) a GIN model on Code2, as in \Cref{sec:main}, but using RMSprop instead of Adam. We choose AdamW and RMSProp, since they are commonly used in image classification \cite{adamW, vit,DBLP:conf/iccv/TouvronCSSJ21} and graph representation learning (GRL) \cite{liu2019hyperbolic, li2019fi,rahman2020training}, respectively.

Due to time constraints, we only report results obtained with one random seed (except for GIN-Code2 with Adam, which we already evaluated across three random seeds in our initial submission, see \Cref{tab:grl_results}). Further, again due to time constraints, for the ResNet-34 task, we do not conduct a hyper-parameter search of SAM's $\rho$ but set it to $\rho=0.05$, as this value has been reported to be a good default value \cite{SAM}.

\Cref{tab:switching_optimizers} show the test performances with switched base optimizers. First, discussing task (i), we note that AdamW under-fits the model and generalizes poorly compared to SGD. Here, SAM exacerbates the performance even further, performing even worse than the AdamW baseline. The reasons for that are unclear, and we leave an investigation into them for future work.  

For task (ii) using RMSprop, we observe that both flat-minima optimizers improve over the baseline performance. However, compared to when using Adam, they perform even more similarly, with SAM only being $0.03\%$ better. Interestingly, the combination WASAM again performs best.

\begin{table}[h!]

    \centering
    \caption{\textbf{Test accuracies/F1 score of switched base optimizers}.}
    \scalebox{1.}{
\begin{tabular}{l|c|c|c|c}
\toprule
    Task & Baseline& SWA & SAM & WASAM \\  \midrule
ResNet34 on CIFAR100 (SGD)& $76.14$ & \good{ + $1.50$}& \good{+$1.91$} & \good{ + $\best{2.60}$}\\
ResNet34 on CIFAR100 (AdamW)& $72.14$ & \good{+ $\best{0.57}$}& \bad{-$2.29$} & \bad{-$1.11$}\\
     \midrule
      GIN on Code2 (Adam) & $15.73_{\pm 0.11}$ &  \good{ + $0.83_{\pm 0.11}$} &  \good{ + $0.57_{\pm 0.09}$} &  \good{ + $\best{1.10_{\pm 0.09}}$} \\
     GIN on Code2 (RMSprop) & $15.30$ & \good{+ $0.67$}& \good{+ $0.70$}& \good{  + $\best{1.62}$} \\ \bottomrule
    \end{tabular}}
    \label{tab:switching_optimizers}
    \vspace{-3ex}
\end{table}

\subsection{Does changing the data augmentation impact SWA's and SAM's effectiveness?}
\label{sec:da}
In this ablation, we want to understand whether different amounts and data augmentation strategies impact SWA's or SAM's effectiveness. As an experimental setup, we consider training a ResNet18 on CIFAR100 for 200 epochs with SGD with a momentum of $0.9$, initial learning rate $0.1$, and cosine learning rate schedule. 
The three data augmentation strategies are (i) none, (ii) basic (random crop, random horizontal flipping), and (iii) AutoAugment following the CIFAR10 policy \cite{cubuk2019autoaugment}. 

\Cref{tab:switching_augmentations} shows the results. We find that with no data augmentation used, SWA, SAM and WASAM improve the baseline results by about the same amount, while SAM improves over SWA when data augmentation is used, and WASAM performs best.

\begin{table}[h!]

    \centering
    \caption{\textbf{Test accuracies of ResNet18 on Cifar100 with different data augmentation schemes}.}
    \scalebox{1.}{
\begin{tabular}{l|c|c|c|c}
\toprule
    Data Augmentation & Baseline& SWA & SAM & WASAM \\  \midrule
None & $60.78$ & \good{ + $2.23$}& \good{+ $2.20$} & \good{ + $\best{2.24}$}\\
Basic & $76.40$ & \good{+ $0.13$}& \good{+ $0.74$} & \good{+ $\best{1.01}$} \\
AA \cite{cubuk2019autoaugment} & $67.59$ & \good{ + $3.03$}& \good{+ $3.35$} & \good{ + $\best{4.17}$} \\
\bottomrule
    \end{tabular}}
    \label{tab:switching_augmentations}
    \vspace{-3ex}
\end{table}

\subsection{Does using a constant learning rate at the end of training improve SWA?} \label{app:constant_lr}
In this ablation, we aim to understand the impact of a constant learning rate at the end of the training, as originally suggested by \cite{SWA}. We follow the same experimental setup as in \Cref{tab:switching_augmentations} and choose the Basic Data Augmentation. At the last 25\% of training (starting from epoch 150), we set the learning rate to $0.05$. We find that this slightly worsens the SWA performance by $-0.66$\% compared to running SWA without changing the learning rate schedule, as explained in \Cref{sec:hparams}.

\section{Experimental details}
\label{app:experimental_details}
\subsection{Computer Vision}
We mostly adopt the hyper-parameter values from \citet{SAM} for WRN-28-10 and PyramidNet-272, from \citet{vit} for ViT, and from \cite{mlp-mixer} for MLP-Mixer models. We average all results across three random seeds.

\subsubsection{Supervised Classification}
We train WideResNets \cite{wrn} with 28 layers and width 10 (WRN28-10) and PyramidNet \cite{pyramidnet} with 110 layers and widening factor $\alpha=272$ (PyramidNet-272) from scratch. The Vision Transformer (ViT) base model with input patch size $16$ (ViT-B/16) and MLP-Mixer base model with input patch size $16$ (MLP-Mixer-B/16) start from pre-trained checkpoints available at \href{https://console.cloud.google.com/storage/vit_models/}{https://console.cloud.google.com/storage/vit\textunderscore models/}. The reason for using pre-trained checkpoints for the ViT and MLP-Mixer models is that, due to their lack of some inductive biases inherent to CNNs, such as translation equivariance and locality, they do not generalize well when trained on insufficient amounts of data \cite{vit}. \Cref{tab:cv_sc_hparams} shows the hyper-parameters for each architecture.

\begin{table}[h]
    \caption{Hyper-parameters for Supervised Classification (SC): CIFAR-\{10, 100\} (\Cref{tab:cv_results})}
    \label{tab:cv_sc_hparams}
    \centering
    \resizebox{.65\textwidth}{!}{
    \begin{tabular}{l|cccc}
    \toprule
    \textbf{Hyper-Parameter} & \textbf{WRN28-10} & \textbf{PyramidNet-272} & \textbf{ViT-B/16} & \textbf{MLP-Mixer-B/16} \\ \midrule
    Base Optimizer & SGD & SGD & SGD & SGD \\
    Batch size & 256 & 256 & 100 & 170\\
    Data augmentation & \multicolumn{4}{c}{Inception-style + Cutout \cite{cutout}} \\
    Dropout rate & \multicolumn{4}{c}{$0.0$} \\
    Epochs & 200 & 200 &  -- & -- \\
    Gradient clipping norm & -- & -- & 1.0 & 1.0 \\
    Learning rate schedule &  \multicolumn{4}{c}{cosine}  \\
    Peak learning rate & $0.1$ & $0.05$ & $0.03$ & $0.03$ \\
    Steps & -- & -- & 12500 & 12500 \\
    SGD Momentum & \multicolumn{4}{c}{$0.9$}  \\
    Warmup steps & -- & -- & 500 & 500 \\
    Weight decay & $5e-4$ & $5e-4$ & $0.0$ & $0.0$ \\
    \midrule
    \multicolumn{5}{c}{\textbf{CIFAR-10}} \\
    \midrule
    SAM $\rho$ & $0.05$& $0.05$ & $0.1$ & $0.02$ \\
    Averaging start $E$ (SWA) & 60\% &60\% & 75\% & 90\%\\
    Averaging start $E$ (WASAM) & 90\% & 75\% & 75\% & 90\% \\
    \midrule
    \multicolumn{5}{c}{\textbf{CIFAR-100}} \\
    \midrule
    SAM $\rho$ & $0.1$& $0.1$ & $0.2$ & $0.05$ \\
    Averaging start $E$ (SWA) & 60\% &60\% & 75\% & 90\%\\
    Averaging start $E$ (WASAM) & 90\% & 75\% & 75\% & 90\% \\
    \bottomrule
    \end{tabular}}
\end{table}

\subsubsection{Self-Supervised Learning} \Cref{tab:cv_ssl_hparams} shows the hyper-parameters for each SSL method. We use implementations from the lightly package, available at \href{https://github.com/lightly-ai/lightly}{https://github.com/lightly-ai/lightly} \cite{lightly}.  

\begin{table}[h]
    \caption{Hyper-parameters for Self-Supervised Learning (SSL): CIFAR-10, ImageNette, results in (\Cref{tab:cv_results})}
    \label{tab:cv_ssl_hparams}
    \centering
    \resizebox{.8\textwidth}{!}{
    \begin{tabular}{l|cccccc}
    \toprule
    \textbf{Hyper-Parameter} & \textbf{MoCo} & \textbf{SimCLR} & \textbf{SimSiam}  & \textbf{BarlowTwins} & \textbf{BYOL} & \textbf{SwaV} \\ \midrule
    Backbone Network & \multicolumn{6}{c}{ResNet-18} \\
    Base Optimizer & \multicolumn{5}{c}{SGD} & Adam \\
    Data augmentation & \multicolumn{5}{c}{SimCLR \cite{simclr}} & Multi-Crop \cite{swav} \\
    Dropout rate & \multicolumn{6}{c}{$0.0$} \\
    Epochs & \multicolumn{6}{c}{800} \\
    Embedding dimensions & \multicolumn{6}{c}{512} \\
    KNN memory bank size & \multicolumn{6}{c}{$4096$} \\
    Learning rate schedule & \multicolumn{6}{c}{cosine} \\
    Peak learning rate & \multicolumn{5}{c}{$6e-2$} & $1e-3$ \\
    SGD Momentum & \multicolumn{5}{c}{$0.9$} & --  \\
    Weight decay & \multicolumn{5}{c}{$5e-4$} & $1e-6$ \\
    \midrule
    \multicolumn{7}{c}{\textbf{CIFAR-10}} \\
    \midrule
    Batch size & \multicolumn{6}{c}{512} \\
    Crop size & \multicolumn{5}{c}{--} & 32 \\
    Gaussian blur & \multicolumn{6}{c}{$0\%$} \\
    SAM $\rho$ & $0.01$ & $0.01$ & $0.01$ & $0.05$ & $0.01$ & $0.05$\\
    Averaging start $E$ (SWA) & 75\% & 90\% & 75\% & 90\% & 60\% & 60\%\\
    Averaging start $E$ (WASAM) & 90\% & 90\% & 90\% & 90\% & 75\% & 90\% \\
    \midrule
    \multicolumn{7}{c}{\textbf{ImageNette}} \\
    \midrule
    Batch size & \multicolumn{6}{c}{256} \\
    Crop size & \multicolumn{5}{c}{--} & 128, 64 \\ 
    Gaussian blur & \multicolumn{6}{c}{$50\%$} \\
    SAM $\rho$ & $0.01$& $0.01$ &$0.02$ & $0.05$ & $0.05$ & $0.01$  \\
    Averaging start $E$ (SWA) & 50\% & 90\% & 75\% & 75\% & 90\% & 50\%  \\
    Averaging start $E$ (WASAM) & 50\% & 50\% & 90\% & 75\% & 90\% & 50\%  \\
    \bottomrule
    \end{tabular}}
\end{table}

\subsection{Natural Language Processing}
For the task of Open Domain Question Answering, we adapt the hyper-parameter values and the 25 retrieved passages for each question from \citealp{izacard2021leveraging}. We report the Exact Match score of FiD-base model on Natural Questions (NQ) and TriviaQA test sets.
For GLUE benchmark, we report Matthew's Corr for CoLA, Pearson correlation coefficient for STSB, and accuracy for the the rest of the datasets. Results are all evaluated on the dev set of GLUE benchmark. We use the RoBERTa-base as our backbone language model, implemented with Huggingface Transformers \cite{wolf2020transformers}. Most of the task-specific hyper-parameter values are adapted from \citealp{aghajanyan2020better}.

\subsection{Graph Representation Learning}
We mostly adapt the hyper-parameter values from \citet{ogb} for GCN \cite{GCN}, SAGE \cite{SAGE}, and GIN \cite{GIN}, from \citet{rp} for  \cite{CP} and ComplEx \cite{Complex}, and from \citet{deepergcn} for DGCN.
Due to high standard errors, we averaged the results of a few tasks more than three times, as mentioned in the following tables. 
\begin{table}[h]
    \caption{Hyper-parameters for NPP tasks, results in \Cref{tab:grl_results}.}
    \label{tab:grl_npp_hparams}
    \centering
    \resizebox{.45\textwidth}{!}{
    \begin{tabular}{l|cc}
    \toprule
    \textbf{Hyper-Parameter} & \textbf{SAGE} & \textbf{DGCN}\\ \midrule
    \multicolumn{3}{c}{\textbf{NPP: OGB-Proteins}} \\
    \midrule
    Aggregation method & Mean & Softmax \\
    Base optimizer & Adam & Adam \\
    Convolution layer & SAGE & DyResGEN\\
    Dropout rate & 0.0 &0.1 \\
    Hidden dimensions & 256 & 64\\
    Learning rate & 0.01 & 0.001 \\
    Normalization layer & -- & Layer norm \\
    Number of epochs & $2000$ & $1000$ \\
    Number of layers & 3 & 112 \\
    Number of random seeds & 5 & 3\\
    Training cluster number & 1 & 15 \\
    Weight decay & 0.0 & 0.0 \\
    \midrule
    SAM $\rho$ & $0.01$ & $0.02$ \\
    Averaging start $E$ (SWA) & 90\% & 90\% \\
    Averaging start $E$ (WASAM) & 90\% & 90\% \\
    \midrule
    \multicolumn{3}{c}{\textbf{NPP: OGB-Products}} \\
    \midrule
    Aggregation method & Mean & Softmax \\
    Base optimizer & Adam & Adam \\
    Batch size & 20000 & -- \\
    Convolution layer & SAGE & Gen\\
    Dropout rate & 0.5 &0.5 \\
    Evaluation cluster number & -- & 8 \\
    Learning rate & 0.01 & 0.001 \\
    Hidden dimensions & 256 & 128 \\
    Normalization layer & -- & Batch norm \\
    Number of epochs & $30$ & $50$ \\
    Number of layers & 3 & 14 \\
    Number of random seeds & 5 & 3\\
    Training cluster number & -- & 10 \\
    Weight decay & 0.0 & 0.0 \\
    \midrule
    SAM $\rho$ & $0.01$ & $0.02$ \\
    Averaging start $E$ (SWA) & 90\% & 60\% \\
    Averaging start $E$ (WASAM) & 75\% & 90\% \\
    \bottomrule
    \end{tabular}}
    \end{table}

\begin{table}[h]
    \caption{Hyper-parameters for GPP: OGB-Code2, results in \Cref{tab:grl_results}.}
    \label{tab:grl_ogb_code_hparams}
    \centering
    \resizebox{.45\textwidth}{!}{
    \begin{tabular}{l|cc}
    \toprule
    \textbf{Hyper-Parameter} & \textbf{GCN} & \textbf{GIN}\\ \midrule
    \multicolumn{3}{c}{\textbf{GPP: OGB-Code2}} \\
    \midrule
    Aggregation method & Mean & Mean \\
    Base optimizer & Adam & Adam \\
    Batch size & 128 & 128 \\
    Convolution layer & GCN & GIN \\
    Dropout rate & 0.0 & 0.0 \\
    Learning rate & 0.001 & 0.001 \\
    Hidden dimensions & 300 & 300\\
    Normalization layer & Batch norm & Batch norm \\
    Number of random seeds & 3 & 3\\
    Number of epochs & $15$ & $30$ \\
    Number of layers & 5 & 5 \\
    Virtual node embeddings & True & True \\
    Vocabulary size & 5000 & 5000 \\
    Weight decay & 0.0 & 0.0 \\
    \midrule
    SAM $\rho$ & $0.2$ & $0.15$ \\
    Averaging start $E$ (SWA) & 50\% & 50\% \\
    Averaging start $E$ (WASAM) & 50\% & 50\% \\
    \bottomrule
    \end{tabular}}
    \end{table}
    
\begin{table}[h]
    \caption{Hyper-parameters for GPP: OGB-Molpcba, results in \Cref{tab:grl_results}.}
    \label{tab:grl_ogb_molpcba_hparams}
    \centering
    \resizebox{.45\textwidth}{!}{
    \begin{tabular}{l|cc}
    \toprule
    \textbf{Hyper-Parameter} & \textbf{GIN} & \textbf{DGCN}\\ \midrule
    \multicolumn{3}{c}{\textbf{GPP: OGB-Molpcba}} \\
    \midrule
    Aggregation method & Mean & Mean \\
    Batch size & 512 & 512 \\
    Base optimizer & Adam & Adam \\
    Convolution layer & GIN & GEN \\
    Dropout rate & 0.0 & 0.2 \\
    Learning rate & 0.001 & 0.001 \\
    Normalization layer & Batch norm & Batch norm \\
    Number of epochs & $100$ & $50$ \\
    Number of layers & 5 & 14 \\
    Number of random seeds & 3 & 3\\
    Hidden dimensions & 300 & 256\\
    Virtual node embeddings & False & True \\
    Weight decay & 0.0 & 0.0 \\
    \midrule
    SAM $\rho$ & $0.01$ & $0.15$ \\
    Averaging start $E$ (SWA) & 90\% & 75\% \\
    Averaging start $E$ (WASAM) & 90\% & 50\% \\
    \bottomrule
    \end{tabular}}
\end{table}

\begin{table}[h]
    \caption{Hyper-parameters for LPP: OGB-Biokg, results in \Cref{tab:grl_results}.}
    \label{tab:grl_ogb_biokg_hparams}
    \centering
    \resizebox{.45\textwidth}{!}{
    \begin{tabular}{l|cc}
    \toprule
    \textbf{Hyper-Parameter} & \textbf{CP} & \textbf{ComplEx}\\ \midrule
    \multicolumn{3}{c}{\textbf{GPP: OGB-Biokg}} \\
    \midrule
    Base optimizer & Adam & Adam \\
    Batch size & 500 & 500 \\
    Learning rate & 0.1 & 0.1 \\
    Number of random seeds & 3 & 3\\
    Number of epochs & $30$ & $50$ \\
    Rank & 1000 & 1000 \\
    Regularizer & N3 & N3 \\
    Weight decay & 0.0 & 0.0 \\
    \midrule
    SAM $\rho$ & $0.1$ & $0.05$ \\
    Averaging start $E$ (SWA) & 50\% & 50\% \\
    Averaging start $E$ (WASAM) & 90\% & 50\% \\
    \bottomrule
    \end{tabular}}
\end{table}

\begin{table}[h]
    \caption{Hyper-parameters for LPP: OGB-Citation2, results in \Cref{tab:grl_results}.}
    \label{tab:grl_ogb_citation2_hparams}
    \centering
    \resizebox{.45\textwidth}{!}{
    \begin{tabular}{l|cc}
    \toprule
    \textbf{Hyper-Parameter} & \textbf{GCN} & \textbf{SAGE}\\ \midrule
    \multicolumn{3}{c}{\textbf{GPP: OGB-Citation2}} \\
    \midrule
    Aggregation method & Mean & Mean \\
    Base optimizer & Adam & Adam \\
    Batch size & 256 & 512 \\
    Convolution layer & GCN & SAGE \\
    Dropout rate & 0.0 & 0.2 \\
    Hidden dimensions & 256 & 256\\
    Number of epochs & $300$ & $300$ \\
    Number of layers & 3 & 3 \\
    Number of random seeds & 3 & 3\\
    Normalization layer & -- & -- \\
    Learning rate & 0.001 & 0.0005 \\
    Virtual node embeddings & False & False \\
    Weight decay & 0.0 & 0.0 \\
    \midrule
    SAM $\rho$ & $0.02$ & $0.01$ \\
    Averaging start $E$ (SWA) & 75\% & 90\% \\
    Averaging start $E$ (WASAM) & 60\% & 90\% \\
    \bottomrule
    \end{tabular}}
\end{table}

\end{document}